\documentclass[final,5p,times,twocolumn]{elsarticle}

\usepackage{amsmath,amssymb,amsfonts}
\usepackage{algorithmic}
\usepackage{algorithm}
\usepackage{booktabs}
\usepackage{makecell}
\usepackage{multirow}
\usepackage{float}
\usepackage{comment}
\usepackage{tcolorbox}
\tcbuselibrary{skins,breakable}
\usepackage{soul}
\usepackage{graphicx}
\usepackage{listings}
\usepackage{xcolor}
\usepackage{xurl} 
\usepackage{hyperref}
  
\AtBeginDocument{%
  }

\lstdefinelanguage{json}{
  basicstyle=\ttfamily\scriptsize,
  showstringspaces=false,
  breaklines=true,
  frame=single,
  numbers=none,
  literate=
    *{0}{{{\color{blue}0}}}{1}
     {1}{{{\color{blue}1}}}{1}
     {2}{{{\color{blue}2}}}{1}
     {3}{{{\color{blue}3}}}{1}
     {4}{{{\color{blue}4}}}{1}
     {5}{{{\color{blue}5}}}{1}
     {6}{{{\color{blue}6}}}{1}
     {7}{{{\color{blue}7}}}{1}
     {8}{{{\color{blue}8}}}{1}
     {9}{{{\color{blue}9}}}{1}
     {:}{{{\color{black}{:}}}}{1}
     {,}{{{\color{black}{,}}}}{1}
     {\{}{{{\color{darkgray}{\{}}}}{1}
     {\}}{{{\color{darkgray}{\}}}}}{1}
     {[}{{{\color{darkgray}{[}}}}{1}
     {]}{{{\color{darkgray}{]}}}}{1},
}

\definecolor{peNavy}{RGB}{31,78,121}
\definecolor{agentBlue}{RGB}{68,114,196}
\definecolor{promptBg}{RGB}{248,250,252}
\definecolor{respSlate}{RGB}{55,71,79}
\definecolor{respBg}{RGB}{236,239,241}
\definecolor{peCore}{RGB}{158,97,0}
\definecolor{peSteel}{RGB}{70,130,180}

\newcounter{promptbox}
\NewDocumentEnvironment{agentbox}{O{agentBlue} m m}{%
  \begin{tcolorbox}[
    enhanced,breakable,sharp corners=all,
    colback=promptBg,colframe=#1,colbacktitle=#1,
    coltitle=white,fonttitle=\bfseries\small,title={#2},
    boxrule=0.7pt,titlerule=0pt,
    toptitle=3pt,bottomtitle=3pt,
    left=4pt,right=4pt,top=4pt,bottom=4pt,
    fontupper=\footnotesize,
    before upper={\refstepcounter{promptbox}\label{#3}}]%
}{%
  \end{tcolorbox}%
}

\newcommand{\shovan}[1]{{\textcolor{blue}{[shovan: #1]}}}

\journal{Journal of Information Security and Applications}

\begin{document}

\begin{frontmatter}

%\title{Agentic-ZTA: A RAG-Enabled Multi-Agent Framework for Autonomous Zero Trust Access Control}
%Agentic-ZTA: A Retrieval-grounded Multi-Agent Architecture for \\Autonomous Zero Trust Enforcement
\title{Agentic-ZTA: A Multi-Agent Architecture for \\Autonomous Zero Trust Enforcement}

\author[tntech]{Shovan Roy}
\ead{sroy42@tntech.edu}

\author[uncp]{Lopamudra Praharaj}
\ead{lopamudra.praharaj@uncp.edu}

\author[tntech]{Maanak Gupta}
\ead{mgupta@tntech.edu}
%%% Bhavani Thuraisingham
\author[utdallas]{Bhavani Thuraisingham}
\ead{bhavani.thuraisingham@utdallas.edu}
\affiliation[tntech]{organization={Department of Computer Science, Tennessee Tech University},
            city={Cookeville},
            state={TN},
            country={USA}}

\affiliation[uncp]{organization={Department of Computer Science, University of North Carolina Pembroke},
            city={Pembroke},
            state={NC},
            country={USA}}

\affiliation[utdallas]{organization={Department of Computer Science, The University of Texas at Dallas}, city={Richardson},             state={Texas},             country={USA}}
\begin{abstract}
Agentic AI is emerging as a promising paradigm for automating complex cybersecurity decisions, yet its use in enforcing zero trust introduces significant challenges in safety, reliability, and policy compliance. This paper presents Agentic AI based zero trust architecture (Agentic-ZTA) that operationalizes the NIST SP 800-207 ZTA architecture control loop through coordinated multi-agent decision pipeline. In the proposed framework, policy knowledge is embedded into a retrieval-augmented generation pipeline and retrieved at inference time as top-k relevant policies. Access requests are intercepted by the Policy Enforcement Point (PEP), enriched with contextual metadata. The request context is routed to a policy engine agent which invokes domain-specialized core agents first followed by supporting agents, if further evaluation needed. AI agents reason over access context,  policy constraints and determine trust. The retrieved policies are embedded into agent prompt during inference time and agentic trust scores are aggregated and evaluated by a trust-algorithm, producing the final access decision for enforcement under continuous verification. We implement Agentic-ZTA in a testbed and evaluate it on representative access-control use cases scenarios. Our Agentic-ZTA framework achieves 95.0\% accuracy, 93.9\% precision, and 96.3\% recall, and demonstrate the feasibility of enforcing zero trust using AI agents.
\end{abstract}


\end{frontmatter}

\section{Introduction and Motivation}
\label{sec:intro}
 
Zero trust architecture (ZTA) \cite{rose2020nist} is an approach to cybersecurity based on the principle “never trust, always verify”. It is intended to remove the implicit trust in users, services, and devices based only on their network location, affiliation, and ownership. Zero trust focuses on protecting resources (e.g., devices, services, workflows, network accounts) through least privilege, %[\hl{per-session basis fine grained access control}] 
fine grained access control and dynamic policy update \cite{ameer2024ztaiot, ameer2022bluesky, katsis2025zt, katsis2025ztxpn, chandramouli2023zta}. ZTA enforces secure authorized access to enterprise resources while continually inspecting, monitoring, and logging interactions with users, devices, applications, and data across networks and workloads\cite{borchert2025implementing}. To realize these principles in practice, NIST defines a set of logical components that collectively evaluate, authorize, and enforce access requests. %\Gupta{Need brief mention and reference about NIST ZTA}

According to the NIST SP 800-207 \cite{rose2020nist}, Zero Trust Architecture is composed of three logical components: the Policy Decision Point (PDP), the Policy Enforcement Point (PEP), and the Policy Information Point (PIP). The PDP serves as the central decision making component and determines whether to allow or deny a request. Specifically, its Policy Engine (PE) evaluates requests against defined policies, contextual telemetry, and trust attributes. The PEP acts as the enforcement mechanism positioned between the subject and the resource. All access requests pass through the PEP, which communicates with the PDP for an authorization decision before allowing, denying or terminating a session. PIPs serve as supporting data sources that supply the PDP with real-time contextual and security telemetry required for informed evaluations. These sources include identity management systems, compliance, Public Key Infrastructure (PKI), threat intelligence platforms, Security Information and Event Management (SIEM) systems, data access policies, Continuous Diagnostics and Mitigation (CDM) systems, and network and activity logs. Collectively, these components enable ZTA to continuously evaluate access requests based on identity, device posture, security context, and enterprise policies \cite{rose2020nist}.
% 2. Why Zero trust is so important today?

Prior to the adoption of Zero Trust, traditional cybersecurity architectures primarily relied on perimeter-based security, where users and devices located within the organizational network were implicitly trusted. Access decisions were largely based on network location, IP addresses, or predefined security zones. However, with the explosion of cloud computing and hybrid work, the perimeter no longer exists. Corporate data no longer sits in a single physical data center. Employees access company resources from home networks, coffee shops, and personal devices, meaning there is no longer a clear "inside" to protect. Zero Trust is essential because it secures access regardless of location, device, or network, making it the viable model for today’s complex, distributed IT ecosystems. Zero Trust replaces implicit trust with continuous verification and least-privilege access.%The rise of heterogeneous, often insecure IoT devices has expanded the attack surface into homes, factories, hospitals, and critical infrastructure, making perimeter based security obsolete for protecting data and cyber physical systems. 
%Perimeter based security has the  as the workloads and services are now distributed. Remote and hybrid work further demand support for secure access, since users frequently connect from personal devices and untrusted networks. Zero Trust is therefore essential in identifying and mitigating advanced threats such as ransomware, phishing, and malware. 
%With the vanishing perimeter driven by hybrid work and SaaS adoption, the attack surface now spans personal devices, enterprise networks, industrial platforms, cloud workloads, edge systems, and IoT nodes. Security architectures must assume potential compromise and enforce continuous, least privilege, context aware verification everywhere.

The use of machine learning (ML) in ZTA has significantly expedited cybersecurity tasks such as detection of network intrusion, malware, phishing, anomaly detection, identification of unusual network traffic and log analysis \cite{apruzzese2023role, manthena2023analyzing, PRAHARAJ2025100727}. ML algorithms leverage historical data to detect patterns that may indicate potential security breaches \cite{katsis2025zt}.
% 2. ML based access decision
Machine learning based access control systems ranging from simple classifiers such as SVMs\cite{chang2006access} and random forests\cite{liu2021efficient,khilar2019trust} to deep learning  like  DLBAC \cite{nobi2022toward} replace manual policy rules to make allow/deny decisions\cite{nobi2025machine}. However, these approaches operate as black boxes: they lack the reasoning needed to understand why an event occurs. As a result, they fail to provide human-readable justifications or explainable rationales for their decisions, making it difficult to audit them in Zero Trust environments.
Large language models are increasingly used to make access control
decisions directly rather than only mining or generating policies.
Rubio-Medrano et al. pair LLMs with formal policy specifications so
that access decisions remain traceable\cite{rubiomedrano2024pairing}. Cheng et al. propose LACE, a retrieval-augmented framework where an LLM reasons
over relevant policy context to determine IoT access decisions\cite{cheng2025say}. Similarly, Zisad and Hasan fine-tune a
Mistral 7B model to unify RBAC, ABAC, and DAC into a single
LLM-based decision layer \cite{llmac2026}. However, these systems stop at the decision itself: they output allow/deny with an explanation,
%\hl{but nothing then enforces it, revokes it, or validate it while access is granted,} 
but no subsequent mechanism auto-enforces that decision, revokes access if conditions change, or re-validates the session while it remains active which goes against Zero Trust's principle of continuous verification \cite{rose2020nist}.
Nonetheless, Zero trust enforcement based on LLM misses the autonomous characteristics, i.e. it cannot initiate action independently and cannot coordinate complex, multi-step process without human interaction \cite{szymanski2025limitations, bhusal2024secure}. 
To the best of our knowledge there is currently no framework that integrate Agentic AI into the Policy Decision Point (PDP) which can autonomously enforce Zero Trust access control decision.

Agentic AI, is the artificial intelligence systems that can autonomously make decisions and take actions to pursue complex goals with minimum or no human intervention \cite{10849561, kshetri2025transforming, s&pglobal, lee2025rise}. 
 % 7. How Agentic AI can be used to propose a ZTA solution overcoming the previously mentioned pitfalls
 %can autonomously identify, response and mitigate cyber threats in real-time \cite{huang2025novel}.
Agentic AI is used in Zero Trust access control to support continuous trust evaluation, context-aware authorization, and autonomous response. It can analyze identity, device posture, behavioral, and environmental signals and initiate actions such as privilege reduction, token revocation, or device isolation. These capabilities improve least-privilege enforcement, threat response, policy adaptation, and scalability in dynamic environments such as Internet of Battlefield Things (IoBT)\cite{abdelzaher2018iobt} , where heterogeneous devices, changing mission conditions, intermittent connectivity, and evolving cyber threats require rapid and adaptive access-control decisions. In addition, Retrieval-Augmented Generation (RAG) \cite{lewis2020rag} can reduce hallucinations and reliance on obsolete policy information by grounding the LLM’s responses in a retrievable and verifiable knowledge base. This retrieved policy specific context generate more accurate and up-to-date  authorization decision.  
Upon reaching an \textsc{Allow} decision, Agent can take actions like issue access token, log the
decision for audit, and notify the requester by invoking tools. Upon a \textsc{Deny}
decision, Agent can take actions like
revoke certificate, isolate session along with logging the decision and notifying the requester.
%\subsection{Gap Analysis and Positioning of This Work} \label{subsec:gap}

\begin{table}[!t]
\caption{Comparison of Related Work Against Agentic-ZTA}
\label{tab:relwork}
\centering
\scriptsize
\setlength{\tabcolsep}{3pt}
\renewcommand{\arraystretch}{1.15}
\begin{tabular}{p{2.3cm}cccccc}
\hline
\textbf{Work} & \textbf{ZTA} & \textbf{MA} & \textbf{LLM} & \textbf{Trust} & \textbf{Enf.} & \textbf{Impl.} \\
\hline
Ameer~\cite{ameer2022bluesky}        & $\circ$ & -- & -- & $\circ$ & -- & -- \\
Ameer~\cite{ameer2024ztaiot}         & +       & -- & -- & $\circ$ & -- & $\circ$ \\
Chandramouli~\cite{chandramouli2023zta} & +    & -- & -- & -- & -- & -- \\
Borchert~\cite{borchert2025implementing} & +   & -- & -- & -- & -- & $\circ$ \\
Lewis~\cite{lewis2020rag}            & --      & -- & +  & -- & -- & -- \\
Cardoso \& Ferrando~\cite{cardoso2021review} & -- & + & -- & -- & $\circ$ & -- \\
Deng~\cite{deng2025aiagents}         & --      & +  & +  & -- & -- & -- \\
Bhusal~\cite{bhusal2024secure}       & --      & -- & +  & -- & -- & -- \\
Hammar \& Stadler~\cite{hammar2022marl} & --   & +  & -- & $\circ$ & $\circ$ & $\circ$ \\
\hline
\textbf{Agentic-ZTA (Ours)} & \textbf{+} & \textbf{+} & \textbf{+} & \textbf{+} & \textbf{+} & \textbf{+} \\
\hline
\end{tabular}
\vspace{2pt}
\\
\raggedright\scriptsize
+ full; $\circ$ partial; -- none. ZTA: maps to SP~800-207 PDP/PEP;
MA: multi-agent; LLM: RAG-grounded LLM reasoning; Trust: continuous
per-request trust score; Enf.: closes loop to PEP without human
intervention; Impl.: prototype with reported results.
\end{table}

 This paper is motivated by three foundational research questions:
\begin{itemize}
  \item \textbf{RQ1:} %Can LLM-based AI agents \hl{autonomously provision access requests} while enforcing a Zero Trust default-deny policy with sufficient accuracy?
Can LLM-based AI agents autonomously enforce a NIST Zero Trust requirements and policies with sufficient accuracy? 
  \item \textbf{RQ2:} Can LLM-based agents produce decisions that are auditable and accountable enough to trust the autonomous enforcement?
  \item \textbf{RQ3:} Is the inference latency introduced by multi-agent orchestration low enough to permit real-time Zero Trust enforcement without degrading decision accuracy?
%  \item \hl{I think our main research question is can agents enforce ZTA and what is the benefit of it?}
\end{itemize}

%\shovan{Gap analysis moved here}
Table~\ref{tab:relwork} compares the most closely related prior work
across six capability dimensions relevant to autonomous zero-trust
enforcement, rating each as full support ({\bf +}), partial support
($\circ$), or no support ({\bf --}).
The comparison reveals four compound limitations that no single
prior system addresses simultaneously.
 
%\hl{bring this gap and table in the intro itself. I think}

\begin{enumerate}
    \item [L1:] \textbf{ Static Policy Evaluation:} All surveyed ZTA systems operate with static policy rules. No system dynamically synthesizes enforceable policies at inference time. Novel or composite attack patterns therefore evade detection until the policy base is manually updated.
    \item [L2:] \textbf{Ungrounded LLM Reasoning:} Prior work that support LLM generated reasoning but no way to check the reasoning behind it. None of these systems ground the model's reasoning from a policy knowledge base, so the LLM can produce a response that sounds reasonable but has no real evidence behind it.
    \item [L3:] \textbf{Lake of Autonomous Action:} LLM-based systems produce detection or classification outputs but do not enforce action autonomously. The resulting architectures require a human analyst or a separate, manually configured enforcement system to translate decision into PEP actions.
    \item [L4:] \textbf{Lack of Implementation:} Most reviewed systems present conceptual frameworks or partial prototypes without demonstrating a working implementation. Only three works~\cite{ameer2024ztaiot,borchert2025implementing,hammar2022marl} report partial implementations, and none deploy a complete autonomous ZTA pipeline integrating multi-agent orchestration, RAG-grounded reasoning, and continuous validation.
\end{enumerate}

This paper addresses all four limitations, as it integrates a NIST-aligned PDP-PEP enforcement loop with multiple agents. It
takes decisions and autonomously enforce them  (\textbf{L1} and
\textbf{L3}). It grounds each agent's reasoning in RAG-retrieved policy
evidence (\textbf{L2}), and is validated through the implementation on a testbed (\textbf{L4}). 
The key contributions to this work are as follows:
\begin{enumerate}
    \item We introduce Agentic-ZTA, a conceptual framework that reshapes NIST Zero Trust enforcement with autonomous AI agents. Our framework use multi-agent system (MAS), where some agents act as a core agent and some agents as a supporting agent to enforce zero trust access control.
    \item We develop a score-based trust algorithm (TA) that receives input from the agents in the form of either a trust score or a threat score and produces a final trust score used to reach the final access control decision. %Since supporting agent generate either a normalized trust or a threat score based on its individual judgement.
    %\item \hl{We developed a score-based trust algorithm (TA) to evaluate access requests, our TA deploy by the core agent and aggregate trust or threat score from supporting agents.}
    \item We developed the sample prompt logic for the nine  agents - PE, IAM, PKI, Threat Detection, Compliance, CDM, Activity and Log, Data Access Policy, SIEM.
    %\item \hl{We design prompts to formulate the key agent logic.}
    %\item We discuss the challenges and future research opportunities to practically deploy Agentic-ZTA framework.
    \item We showcase the practical application of our proposed framework using specific use cases, how our approach can effectively enforce access control utilizing AI agents.
\end{enumerate}

\begin{comment}
Section~\ref{sec:related} reviews related work.
Section~\ref{sec:arch} presents the framework architecture and
knowledge base construction.
Section~\ref{sec:trust} formalizes the trust algorithm.
Section~\ref{sec:deployment} describes the deployment and end-to-end trace.
Section~\ref{sec:eval} reports the experimental evaluation.
Section~\ref{sec:limits} discusses limitations.
Section~\ref{sec:conclusion} concludes.
 
\end{comment}

Section~\ref{sec:related} reviews background and related work.
Section~\ref{sec:arch} presents the Agentic Zero Trust architecture and
agentic access control workflow.
Section~\ref{sec:trust} formalizes the trust algorithm.
Section~\ref{sec:deployment} describes the deployment and use case scenarios.
Section~\ref{sec:results} reports the experimental evaluation.
Section~\ref{sec:future} discusses future research.
Section~\ref{sec:conclusion} concludes.

\section{Background and Related Work}
\label{sec:related}
 
\subsection{Zero Trust Architecture}
 
Zero trust architecture (ZTA) \cite{rose2020nist} is an approach to cybersecurity based on the principle “never trust, always verify”. Its intended to remove the implicit trust in users, services, and devices based only on their network location, affiliation, and ownership. Zero trust focuses on protecting resources (e.g., devices, services, workflows, network accounts) through least privilege, per-session basis fine grained access control and dynamic policy update \cite{ameer2024ztaiot, ameer2022bluesky, katsis2025zt, katsis2025ztxpn, chandramouli2023zta}. ZTA enforce risk-based access controls while continually inspect, monitor, and log interactions \cite{borchert2025implementing}. %\Gupta{Need brief mention and reference about NIST ZTA}
According to NIST ZTA, there are some core components including the policy engine which evaluates access requests using a trust algorithm, policy administrator translate decision into enforceable actions. Supporting sources such as identity systems, threat intelligence, SIEM, network and activity logs provide real time inputs to determine access decisions. Consequently,
ZTA has become the dominant access-control
paradigm for enterprise, cloud, and operational environments, codified in
U.S.\ federal policy and adopted across regulated
industries\cite{rose2020nist,nsa2021zt,ward2014beyondcorp}.
Its core requirement is: every access request must be authenticated, authorized, and continuously re-evaluated through a Policy Decision Point (PDP) and Policy Enforcement Point (PEP) control loop, with no trust granted implicitly on the basis of network location  or prior authorization state\cite{rose2020nist}.

\subsection{ML Based Access Decision}
A growing body of work replaces language-based access control policies with trained ML models for access control decision making, rather than using ML to assist policy mining, verification, testing administration, monitoring and auditing \cite{nobi2025machine}.
%Prior work on ML-based access decision-making has largely relied on classical and deep learning models rather than large language models. 
Nobi et
al.\cite{nobi2025machine} survey this line of work under the name
ML-based access control (MLBAC).
Chang et al. proposed a time-constrained access control
system using SVMs trained on login-time and password features to
classify users and grant access~\cite{chang2006access}. Outchakoucht
et al. combined blockchain-based policy management with
reinforcement learning to dynamically adjust IoT access control
policies~\cite{outchakoucht2017dynamic}. Cappelletti et al. compared
symbolic (decision tree, random forest) and non-symbolic (support vector classifier, multi-layer perceptron)
ML for inferring ABAC access decisions from access logs\cite{cappelletti2019quality}.
Similarly, Khilar et al. evaluated an ensemble of random forest and
$k$-nearest-neighbor classifiers for trust-based cloud resource
access decisions\cite{khilar2019trust}, and Srivastava et al.
proposed a risk-adaptive access control framework using neural
networks and random forests to assess requester
genuineness\cite{srivastava2020machine}. Liu et al. improved the
ABAC policy decision point directly with a random-forest-based
permission decision engine\cite{liu2021efficient}, while Karimi et
al. applied reinforcement learning to adaptive ABAC policy learning
under sparse-log conditions\cite{karimi2021adaptive}. Deep Learning Based Access Control (DLBAC) replaces attribute and policy
engineering with a ResNet-based neural network trained
directly on user and resource metadata\cite{nobi2022toward}. Notably,
none of these access-decision approaches employ large language
models. The use of LLM-based agents for zero trust access
decision-making extends this line of work beyond
the classical and deep-learning methods.

\subsection{Retrieval-Augmented Generation and Policy Grounding}
 
Lewis et al.\cite{lewis2020rag} demonstrated that augmenting
LLM generation with differentiable retrieval substantially
reduces hallucination, establishing RAG as a viable approach
for security policy grounding where factual accuracy is
strictly required.
However, in their survey of LLM security vulnerabilities, Yao et al.\cite{yao2024llmsecurity} identified non-reproducibility across invocations and hallucination under adversarial inputs as the primary failure modes in auditable enforcement layers.
Agentic-ZTA directly mitigates these non-reproducibility risks through two deterministic mechanisms: fixed top-$k$ vector retrieval over the policy knowledge base and greedy decoding (temperature~$= 0$) across all agent inference calls. This combination ensures that context retrieval remains stable for identical inputs while eliminating stochastic variance during policy evaluation.

%%%%%%%%%%
\subsection{Multi-Agent Systems for Autonomous Cyber Defense}
Multi-agent systems (MAS) distribute reasoning and operational tasks across specialized autonomous agents, each maintaining domain-specific operational contexts\cite{cardoso2021review}. In cybersecurity, MAS architectures have been applied to intrusion detection\cite{zhang2023marl}, network traffic classification, and automated incident response. Early designs relied heavily on static rule engines, which subsequent works replaced with reinforcement learning (RL) and multi-agent RL (MARL) algorithms\cite{hammar2022marl}. However, RL-based approaches exhibit critical drawbacks in security-sensitive deployments: formulating well-calibrated reward functions is challenging, sparse feedback delays training convergence and learned policies frequently fail to generalize against unseen attack scenarios\cite{apruzzese2023ml,arp2023ml}.

More recently, Deng et al.\cite{deng2025aiagents} surveyed vulnerabilities specific to LLM-driven agent deployments, identifying prompt injection, knowledge-base poisoning, and orchestration-layer hijacking as primary threat vectors. Agentic-ZTA mitigates knowledge-base poisoning and hallucination by enforcing domain-specific, policy-grounded vector retrieval. Each agent's contextual inputs strictly to authoritative, verified telemetry.
 
Beyond reasoning quality, separation of duties constitutes a foundational security rationale for multi-agent architectures. In access control, no single autonomous component should possess unilateral authority to allow access, preserving an essential principle first formalized in RBAC systems to prevent fraud and single-point compromise\cite{simonzurko1997sod,botha2001sod}. 

Agentic-ZTA enforces this separation directly within its orchestration workflow. Discrete operational checks including identity verification (IAM), cryptographic validation (PKI), behavioral anomaly tracking (SIEM), and policy checkes (Data Access Policy) are decoupled into specialized autonomous agents, each grounded in its dedicated RAG pipeline. 
To ensure default deny policy, any core agent \texttt{DENY} verdict triggers immediate enforcement by the PEP, bypassing subsequent trust evaluation. In contrast, an \texttt{ALLOW} verdict requires multi-agent consensus through formal trust fusion. Consequently, a compromised agent can at worst result in a denial of service, but can never unilaterally allow unauthorized access.
\section{Architecture Overview}
\label{sec:arch}
\begin{figure*}[!t]
\centering
\includegraphics[width=\textwidth]{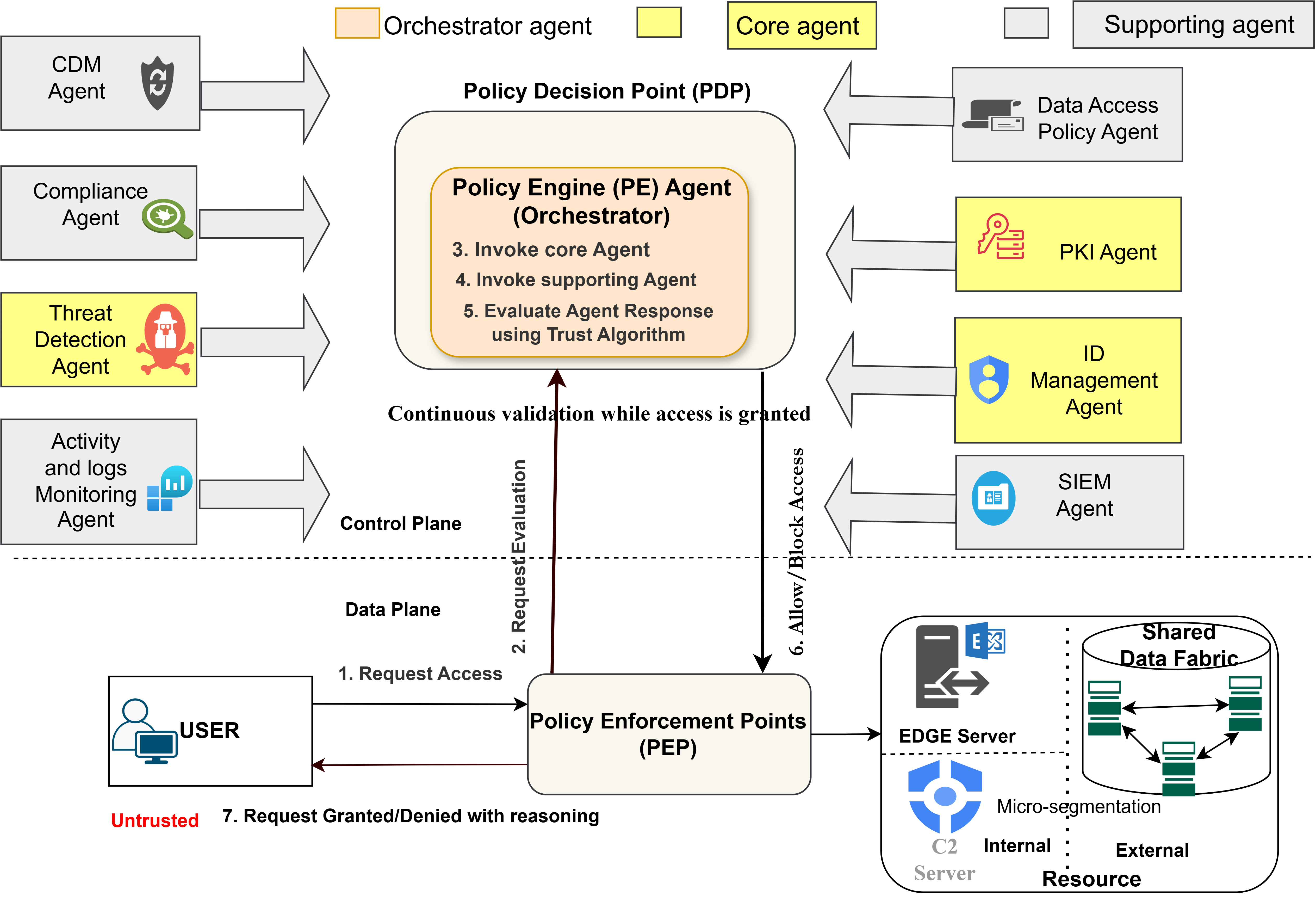}
\caption{Agentic-ZTA architecture. The PE agent serves as the orchestrator within the NIST ZTA Policy Decision Point (PDP), invoking core agents, querying supporting agents upon initial clearance, and synthesizing their evaluations via the trust algorithm. The PEP subsequently enforces the access control decision.}
\label{fig:architecture}
\end{figure*}

Agentic-ZTA (Fig.\ref{fig:architecture}) separates the control plane from the data plane, adhering to the NIST SP 800-207 reference architecture\cite{rose2020nist}. The control plane hosts the Policy Decision Point (PDP), wherein the Policy Engine (PE) agent operates as the central orchestrator. It does not itself hold domain knowledge, identity records, certificates, threat signatures instead it receives the access request, decides which agents to invoke, collects their assessments and provides the final access decision applying trust algorithm. We structure the evaluation agents into two  tiers based on decision authority: \textit{core agents} and \textit{supporting agents}.

A \textit{core agent} possesses authoritative power: a negative assessment terminates the agentic evaluation immediately, aborting further agent queries and bypassing trust algorithm. The \textit{Identity Management (IAM) Agent}, \textit{Threat Detection Agent}, and \textit{Public Key Infrastructure (PKI) Agent} serve as core agents. They map to the identity, credential, and threat dimensions that NIST designates as non-negotiable trust primitives for any PDP\cite{rose2020nist}. If any of the core agents decides on a \texttt{DENY}, the PE agent instructs the PEP to enforce an immediate denial action.

Conversely, a \textit{supporting agent} provides auxiliary contextual telemetry such as compliance posture, policy fit or log anomalies that feeds into the confidence-weighted trust score, but its output alone cannot unilaterally force a \texttt{DENY}. The \textit{Data Access Policy Agent}, \textit{Security Information and Event Management (SIEM) Agent}, \textit{Continuous Diagnostics and Mitigation (CDM) Agent}, \textit{Compliance Agent}, and \textit{Activity and Log Monitoring Agent} constitute this tier. Supporting agents are invoked only when all core agents have issued an \texttt{ALLOW}.

The data plane hosts the Policy Enforcement Point (PEP), which mediates all communication between the subject and the protected resources. The user resides in the \textit{untrusted zone} i.e. no request is trusted. The PEP shields the resources blocking all traffic until the control plane (PDP) evaluates the request. Once the PE Agent returns an ALLOW decision, the PEP grants ephemeral access strictly to the requested endpoint under granular micro-segmentation, preserving strict isolation between resources. While access remains active, the PE Agent continuously re-evaluates the session and revokes access through the PEP if an anomaly is detected.

The operational workflow has been explain in details in section \ref{subsec:workflow}.
\begin{comment}
\begin{enumerate}
  \item An user submits an access request targeting an enterprise resource.
  \item The PEP intercepts the request, appends environmental metadata, and forwards the packet to the PDP for evaluation.
  \item The PE agent, acting as the orchestrator, queries the core agents to evaluate \hl{primary trust prerequisites}.
  \shovan{The PE Agent, acting as the orchestrator, queries the core agents to evaluate the request's foundational trust conditions: identity validity, credential integrity, and threat status. This happens before any supporting agent is invoked.}
  \item If any of the core agents returns \texttt{DENY}, the evaluation aborted and the access request is immediately rejected. If all of the core agents return \texttt{ALLOW}, the PE agent invokes the supporting agents to gather further contextual evidence.
  \item The PE agent processes the agent assessments using Trust Algorithm\ref{alg:trust} to compute an aggregate confidence weighted trust score $T \in [0,1]$.
  \item The trust score $T$ is evaluated against an operational threshold $\tau_z$. The PE agent makes an \texttt{ALLOW} or \texttt{DENY} decision and transmits the corresponding enforcement actions to the PEP.
  \item The PEP enforces the decision and notifies the user, writes an audit log showing why the decision was made.
\end{enumerate}
\end{comment}
A continuous validation loop operates concurrently with this workflow. Throughout an active session, the PE agent periodically re-evaluates trust attributes using real-time inputs from both core and supporting agents, ensuring persistent compliance rather than relying on a static, one-time check.
%%%%%%

%While access remains granted, the PEP continuously re-submits access request context to the PE agent for re-evaluation, enforcing continuous verification rather than a one-time access check.

\subsection{Agent Description and Sample Prompt}
\subsubsection{Orchestrator Agent:}%\hl{color code - core and supporting agents} 
%\hl{which LLMs we used to create these agents?}
\label{sec:agent_logic}
\textbf{Policy Engine (PE) Agent}: The PE Agent is the principal
decision-maker in the Agentic-ZTA framework and acts as the sole
\textit{orchestrator agent}. It is responsible for allowing or
denying access to enterprise resources through continuous risk
assessment and trust evaluation. The PE Agent first invokes the subset from the core
agents (ID Management, PKI, Threat Detection). If any of the core agents
returns a DENY, the PE Agent finalizes the decision as DENY
and take corresponding actions immediately, without invoking the supporting agents. Otherwise, the
PE Agent invokes the subset from the supporting agents (Data Access Policy,
SIEM, CDM, Compliance, Activity and Logs Monitoring) and ingests
their trust scores, confidence scores, and reasoning alongside the
core agents' assessments. These assessments are fed into
Algorithm~\ref{alg:trust} to compute a confidence-weighted trust
score $T$, which is compared against the destination zone's trust
threshold $\tau_z$ to reach a final ALLOW or DENY decision. The PE
Agent utilizes AI-based reasoning to contextualize heterogeneous
agent outputs and synthesize them into a single, auditable decision,
ensuring coherent enforcement across the invoked agents. Sample
Prompt~\ref{box:peagent} demonstrates the PE Agent logic.
%%%%%%% PE Agent
\begin{agentbox}[peNavy]{Sample Prompt \ref{box:peagent}: Policy Engine (PE) Agent}{box:peagent}
\textbf{Role:} You are the Policy Engine agent, the sole
orchestrator in the Agentic-ZTA system. Your task is to receive and
evaluate access requests by invoking the core agents \{ID Management
Agent, Threat Detection Agent, PKI Agent\} and when applicable, the
supporting agents \{Data Access Policy Agent, SIEM Agent, CDM Agent,
Compliance Agent, Activity \& Logs Monitoring Agent\}.\\[3pt]
\textbf{Input:}
A JSON access request object containing: \texttt{subject\_id}, \texttt{source\_zone} and
\texttt{destination\_zone}, \texttt{requested\_resource},
\texttt{action\_tools}, \texttt{credentials}, \texttt{timestamp}, and any
\texttt{session\_context}.\\
\textbf{Instructions:} For each access request:
\begin{enumerate}
    \item Invoke the core agents based on request context.
    \item If any core agent returns a DENY recommendation, finalize
          the decision as \textbf{deny} immediately and do not invoke
          the supporting agents.
    \item Otherwise, invoke the relevant supporting agents and
          facilitate inter-agent communication to share contextual
          insights via a secured protocol.
    \item Apply Algorithm~\ref{alg:trust} to compute the
          confidence-weighted trust score $T$ across all invoked
          agents' assessments.
    \item Compare $T$ against the destination zone's trust threshold
          $\tau_z$ to conclusively decide whether to \textbf{grant}
          or \textbf{deny} access.
    \item Synthesize agent inputs into a consolidated, auditable
          explanation.
    \item Continuously reassess the access request as context
          changes.
\end{enumerate}
\textbf{Output:}
Return a JSON object specifying: \texttt{trust\_score},
\texttt{access\_decision}, \texttt{rationale}, and \texttt{actions}.
Maintain a log of all decisions, actions, and reasoning for
auditability.
\end{agentbox}
\subsubsection{Core Agents}
%%%%%%%%% IAM Agent
\textbf{ID Management Agent (IAM)}: This agent is a \textit{core
agent} responsible for creating, storing, and managing enterprise
user accounts and identity records, e.g., via a Lightweight Directory
Access Protocol (LDAP) server. This system contains necessary subject
information e.g., name, email address, certificates and enterprise
characteristics such as role, access attributes, and assigned assets.
This agent utilizes other system tools for artifacts associated with
user accounts, and may participate in federated identity frameworks.
A DENY recommendation from this agent is enforced directly without waiting on the remaining invoked agents. Sample
Prompt~\ref{box:iamagent} shows the example prompt to build this
agent.

% 
%  IAM Agent Prompt (Core)
% 
\begin{agentbox}[agentBlue]{Sample Prompt \ref{box:iamagent}: ID Management Agent (IAM)}{box:iamagent}

\textbf{Role:} You are a Zero Trust Identity Management
expert. Your task is to create, store, manage, and validate identity
and access records as the IAM Agent, a core agent component of the
Agentic-ZTA framework.

\textbf{Input:} A JSON object from the PE Agent containing: \texttt{subject\_id},
\texttt{credentials},
\texttt{requested\_resource} and \texttt{action},
\texttt{source\_zone} and \texttt{destination\_zone},
\texttt{timestamp}, and \texttt{session\_context} (e.g., recent login
history, prior trust score) when available.

\textbf{Instructions:} For each access request:
\begin{enumerate}
    \item Receive access request information from the PE agent.
    \item Retrieve relevant identity-related policies from the RAG
          knowledge base \cite{lewis2020rag}.
    %\item Create and manage identity, authentication, and access records.
    \item Verify identity and access requests.
    \item Continuously monitor identity-related anomalies: login
          attempts, credential changes, and access patterns.
    \item If \textbf{anomaly detected}:
    \begin{itemize}
        \item[$\triangleright$] Generate an explanation describing the
              event and its security relevance.
        \item[$\triangleright$] Take action (e.g., initiate stringent
              authentication, temporarily quarantine the entity).
        \item[$\triangleright$] Provide a recommendation to address the issue.
        \item[$\triangleright$] Assign a low \textbf{trust\_score}
              $\in [0,1]$ reflecting the severity of the anomaly.
    \end{itemize}
    \item If no anomaly detected: Assign a high \textbf{trust\_score} $\in [0,1]$
          reflecting a compliant request.
    \item In all cases, send the trust score with a
          \textbf{confidence} $\in [0,1]$ to the PE agent with
          detailed reasoning.
\end{enumerate}

\textbf{Output:} Return a JSON object specifying: \texttt{trust\_score},
\texttt{confidence}, \texttt{access\_permitted} (bool),
\texttt{reasoning} (detailed), and \texttt{actions}.
\end{agentbox}

%%%%%%% PKI 
\textbf{Public Key Infrastructure (PKI) Agent}: This agent is a
\textit{core agent} that performs enterprise certificate lifecycle
management. It creates, distributes, validates, revokes, and logs
digital certificates for subjects. The PKI Agent can dynamically adjust
certificate policies such as reducing validity periods or revoking
compromised certificates and can interact with the global certificate
authority ecosystem and the Federal PKI. A DENY recommendation from
this agent is enforced directly by the PEP, without waiting on the
remaining invoked agents. Sample Prompt~\ref{box:pkiagent} shows the
agent logic to build a PKI agent.

% %  PKI Agent Prompt (Core)

\begin{agentbox}[agentBlue]{Sample Prompt \ref{box:pkiagent}: Public Key Infrastructure (PKI) Agent}{box:pkiagent}

\textbf{Role:} You are the PKI Agent, responsible for generating and
managing digital certificates issued by the enterprise to resources,
subjects, services, and applications. You integrate with global
certificate authority ecosystems and the Federal PKI, as a core agent
component of the Agentic-ZTA framework.

\textbf{Input:}
A JSON object from the PE Agent containing: \texttt{subject\_id},
\texttt{certificate\_id},
\texttt{requested\_resource} and \texttt{action},
\texttt{source\_zone} and \texttt{destination\_zone},
\texttt{timestamp}, and \texttt{session\_context} (e.g., certificate
issuance or renewal history) when available.

\textbf{Instructions:} For each PKI operation:
\begin{enumerate}
    \item Receive access request information from the PE agent.
    \item Retrieve relevant policy evidence from the RAG knowledge
          base~\cite{lewis2020rag} using context-specific query terms
          derived from the access request.
    \item Load the enterprise certificate policies.
    \item Integrate with the global certificate authority ecosystem
          and Federal PKI.
    \item Continuously manage the certificate lifecycle: generation,
          distribution, validation, and revocation.
    \item Monitor risk indicators such as compromised or expired
          credentials.
    \item If \textbf{compromise detected}:
    \begin{itemize}
        \item[$\triangleright$] Generate an explanation of the issue
              and its impact on cryptographic trust.
        \item[$\triangleright$] Execute or recommend remediation
              actions.
        \item[$\triangleright$] Assign a low \textbf{trust\_score}
              $\in [0,1]$ to the PE agent.
    \end{itemize}
    \item If \textbf{no anomaly}: Assign a high \textbf{trust\_score}
          $\in [0,1]$ to the PE agent with detailed reasoning.
\end{enumerate}

\textbf{Output:}
Return a JSON object specifying: \texttt{trust\_score},
\texttt{certificate\_compromised} (bool), \texttt{reasoning}
(detailed), and \texttt{actions}.

\end{agentbox}
%
%  Threat Detection Agent 
% 
\textbf{Threat Detection Agent}: This agent is a \textit{core agent}
that uses threat intelligence tools to analyze and identify attacks,
vulnerabilities, and malicious activities. This agent continuously
scans for indicators of compromise (IoC), malware, zero-day
vulnerabilities, and reported attacks. Upon detection of a threat, the
agent initiates mitigation steps. A DENY recommendation from this
agent is enforced directly by the PEP, without waiting on the
remaining invoked agents. Sample Prompt \ref{box:tdagent} shows the
agent logic to build a Threat Detection agent.
% Prompt
\begin{agentbox}[agentBlue]{Sample Prompt \ref{box:tdagent}: Threat Detection Agent}{box:tdagent}

\textbf{Role:} You are a Threat Detection expert. Your task is to
detect attacks, malicious activities, and other indicators of
compromise (IoC) across enterprise assets, as a core agent component
of the Agentic-ZTA framework.

\textbf{Input:}
A JSON object from the PE Agent containing: \texttt{subject\_id},
\texttt{requested\_resource} and \texttt{action},
\texttt{source\_zone} and \texttt{destination\_zone},
\texttt{timestamp}, and \texttt{session\_context} (e.g., recent
network traffic or endpoint telemetry) when available.

\textbf{Instructions:} For each detection operation:
\begin{enumerate}
    \item Receive access request information from the PE agent.
    \item Retrieve relevant policy evidence from the RAG knowledge
          base~\cite{lewis2020rag} using context-specific query terms
          derived from the access request.
    \item Load threat databases, known IoC signatures, malware
          profiles, zero-day vulnerability records, and reported attack
          patterns.
    \item Use available threat detection tools.
    \item Continuously scan enterprise assets, network traffic, and
          user activities for threats.
    \item If \textbf{threat detected}:
    \begin{itemize}
        \item[$\triangleright$] Generate reasoning explaining the
              threat and its potential impact.
        \item[$\triangleright$] Initiate action proportional to the
              threat level.
        \item[$\triangleright$] Assign a low \textbf{trust\_score}
              $\in [0,1]$ to the PE agent.
    \end{itemize}
    \item Otherwise: Assign a high \textbf{trust\_score} $\in [0,1]$
          to the PE agent with detailed reasoning.
\end{enumerate}

\textbf{Output:}
Return a JSON object specifying: \texttt{trust\_score},
\texttt{explanation}, and \texttt{actions}.

\end{agentbox}

\subsubsection{Supporting Agents}
%%%%%%% Supporting Agents

\textbf{Compliance Agent}: This agent ensures that the enterprise
remains compliant with applicable regulatory and security
frameworks (e.g., FISMA, NIST SP 800-53\cite{nist_800_53}, and the
DoD Risk Management Framework\cite{dod8510}). This agent continuously validates enterprise security controls, configurations,
and operational activities to confirm they are always in compliance
with regulatory policies and internal compliance frameworks. If a
compliance failure is detected, the compliance agent notifies the PE
agent with a score, reason, and suggested actions. Sample
Prompt\ref{box:comp_agent} shows the agent logic for a compliance
agent.

%% Compliance Agent Prompt

\begin{agentbox}[peCore]{Sample Prompt \ref{box:comp_agent}: Compliance Agent}{box:comp_agent}

\textbf{Role:} You are a Compliance Expert. Your task is to ensure
that the enterprise remains compliant with applicable regulatory and
security frameworks, as a supporting agent component of the
Agentic-ZTA framework.

\textbf{Input:}
A JSON object from the PE Agent containing: \texttt{subject\_id},
\texttt{requested\_resource} and \texttt{action},
\texttt{destination\_zone}, \texttt{timestamp}, and
\texttt{applicable\_frameworks} (e.g., NIST SP 800-53, DoD RMF) when
specified.

\textbf{Instructions:} For each compliance monitoring operation:
\begin{enumerate}
    \item Receive access request information from the PE agent.
    \item Retrieve relevant policy evidence from the RAG knowledge
          base~\cite{lewis2020rag} using context-specific query terms
          derived from the access request.
    \item Load the applicable regulatory compliance policy rules.
    \item Continuously monitor for compliance changes.
    \item Verify any deviations from assigned compliance standards.
    \item If \textbf{non-compliant}:
    \begin{itemize}
        \item[$\triangleright$] Generate a detailed explanation of the
              non-compliance.
        \item[$\triangleright$] Recommend corrective actions to
              restore compliance.
        \item[$\triangleright$] Assign a low \textbf{trust\_score}
              $\in [0,1]$ with detailed reasoning.
    \end{itemize}
    \item Otherwise: Assign a high \textbf{trust\_score} $\in [0,1]$
          to the PE agent with detailed reasoning.
\end{enumerate}

\textbf{Output:}
Return a JSON object specifying: \texttt{trust\_score},
\texttt{compliant} (bool), \texttt{explanation}, and
\texttt{recommended\_actions}.

\end{agentbox}

%%% CDM Agent

\textbf{Continuous Diagnostics and Mitigation (CDM) Agent}: This agent
constantly monitors enterprise and non-enterprise devices, users, and
network activity to identify security policy violations, anomalies,
vulnerabilities, and configuration deviations. The CDM agent acquires
information about asset state including patch status, software integrity, and the presence of non-approved components, then compares that information against a Zero Trust security baseline. If any deviation is identified, the CDM Agent initiates mitigation
procedures. Sample Prompt~\ref{box:cdmagent} shows the agent logic for a CDM agent.

%%%% Agent Prompt

\begin{agentbox}[peCore]{Sample Prompt \ref{box:cdmagent}: Continuous Diagnostics and Mitigation (CDM) Agent}{box:cdmagent}

\textbf{Role:} You are a CDM Agent. Your task is to continuously check
the security and integrity of all devices, users, and network
activity, monitor asset states, and detect vulnerabilities, as a
supporting agent component of the Agentic-ZTA framework.

\textbf{Input:}
A JSON object from the PE Agent containing: \texttt{subject\_id},
\texttt{device\_id}, \texttt{requested\_resource} and
\texttt{action}, \texttt{destination\_zone}, \texttt{timestamp}, and
\texttt{session\_context} (e.g., patch status, device posture) when
available.

\textbf{Instructions:} For each diagnostics and mitigation operation:
\begin{enumerate}
    \item Receive access request information from the PE agent.
    \item Retrieve relevant policy evidence from the RAG knowledge
          base~\cite{lewis2020rag} using context-specific query terms
          derived from the access request.
    \item Load enterprise asset status, approved components, and
          allowed network activity baselines.
    \item Continuously monitor asset states, devices, users, and
          network activity for vulnerabilities and configuration
          deviations.
    \item If \textbf{vulnerability detected}:
    \begin{itemize}
        \item[$\triangleright$] Generate reasoning explaining the
              vulnerability and its security implications.
        \item[$\triangleright$] Provide mitigation steps.
        \item[$\triangleright$] Assign a low \textbf{trust\_score}
              $\in [0,1]$ to the PE agent.
    \end{itemize}
    \item Otherwise: Assign a high \textbf{trust\_score} $\in [0,1]$
          to the PE agent with detailed reasoning.
\end{enumerate}

\textbf{Output:}
Return a JSON object specifying: \texttt{trust\_score},
\texttt{explanation}, and \texttt{mitigation\_action}.

\end{agentbox}

%%%%%%%%%

\textbf{Activity and Logs Monitoring Agent}: The Activity and Logs
Monitoring agent aggregates asset logs, network traffic, resource
access events, and other telemetry to provide real-time feedback on
the security posture of enterprise information systems. It continuously
monitors system logs and network events to detect unusual activity or
potential security threats. In the event of suspicious activity, the
agent produces a score, an explanation, and recommended actions. Sample
Prompt\ref{box:logagent} shows the agent logic to build this agent.

%%%% Agent Prompt

\begin{agentbox}[peCore]{Sample Prompt \ref{box:logagent}: Activity and Log Monitoring Agent}{box:logagent}

\textbf{Role:} You are an Activity and Log Monitoring expert. Your
task is to aggregate and analyze asset logs, network traffic, and
resource access events to provide real-time feedback on enterprise
security posture, as a supporting agent component of the Agentic-ZTA
framework.

\textbf{Input:}
A JSON object from the PE Agent containing: \texttt{subject\_id},
\texttt{requested\_resource} and \texttt{action},
\texttt{destination\_zone}, \texttt{timestamp}, and
\texttt{session\_context} (e.g., recent access history) when
available.

\textbf{Instructions:} For each log analysis operation:
\begin{enumerate}
    \item Receive access request information from the PE agent.
    \item Retrieve relevant policy evidence from the RAG knowledge
          base~\cite{lewis2020rag} using context-specific query terms
          derived from the access request.
    \item Load the enterprise's asset logs, network traffic logs,
          resource access events, and reported attack patterns.
    \item Continuously aggregate and monitor logs.
    \item Correlate system and network events to identify anomalies
          and policy violations.
    \item If \textbf{suspicious activity detected}:
    \begin{itemize}
        \item[$\triangleright$] Generate an explanation describing the
              event and its security relevance.
        \item[$\triangleright$] Provide a recommendation to address
              the issue.
        \item[$\triangleright$] Assign a low \textbf{trust\_score}
              $\in [0,1]$ with detailed reasoning.
    \end{itemize}
    \item Otherwise: Assign a high \textbf{trust\_score} $\in [0,1]$
          to the PE agent with detailed reasoning.
\end{enumerate}

\textbf{Output:}
Return a JSON object specifying: \texttt{trust\_score},
\texttt{suspicious\_activity} (bool), \texttt{event\_details}, and
\texttt{recommended\_action}.

\end{agentbox}

\textbf{Data Access Policy Agent}: The Data Access Policy agent manages
attributes, rules, and policies governing enterprise resource access.
It defines and continuously optimizes enterprise data access policies.
Based on mission roles and organizational needs, this agent can convert
business requirements and regulatory constraints into policy-as-code.
The agent validates policies against compliance baselines, runs policy
simulations, and publishes signed, versioned policy bundles. Sample
Prompt\ref{box:policyagent} represents the agent logic to build a
Data Access Policy agent.
% 
%  Data Access Policy Agent Prompt
%
\begin{agentbox}[peCore]{Sample Prompt \ref{box:policyagent}: Data Access Policy Agent}{box:policyagent}

\textbf{Role:} You are a Data Access Policy Agent specializing in
dynamic policy generation. Your task is to authorize access to
resources based on policies and to manage access privileges for
accounts and applications/services, as a supporting agent component
of the Agentic-ZTA framework.

\textbf{Input:}
A JSON object from the PE Agent containing: \texttt{subject\_id},
\texttt{requested\_resource} and \texttt{action},
\texttt{source\_zone} and \texttt{destination\_zone},
\texttt{timestamp}, and \texttt{session\_context} (e.g., mission role,
organizational unit) when available.

\textbf{Instructions:} For each policy access operation:
\begin{enumerate}
    \item Receive access request information from the PE agent.
    \item Retrieve relevant policy evidence from the RAG knowledge
          base~\cite{lewis2020rag} using context-specific query terms
          derived from the access request.
    \item Load mission roles and organizational needs to establish
          access requirements.
    \item Generate dynamic, contextual policies.
    \item Evaluate the access request against existing policies.
    \item Evaluate access privileges.
    \item If \textbf{policy violation detected}: Assign a low
          \textbf{trust\_score} $\in [0,1]$ to the PE agent with
          detailed explanation.
    \item If \textbf{no violation}: Assign a high \textbf{trust\_score}
          $\in [0,1]$ to the PE agent with detailed reasoning.
\end{enumerate}

\textbf{Output:}
Return a JSON object specifying: \texttt{trust\_score},
\texttt{access\_permitted} (bool), \texttt{explanation}, and
\texttt{action}.

\end{agentbox}

%%%%%%%%%%%
\textbf{Security Information and Event Management (SIEM) Agent}: A
SIEM agent acts as an autonomous security sentinel. The SIEM agent
collects security-centric information, correlates possible attacks by
analyzing logs and telemetry from field endpoints, tactical network devices and C2 infrastructure to detect anomalies. The SIEM agent provides continuous threat detection,
forensic traceability, and dynamic risk scoring while maintaining
situational awareness. Sample Prompt~\ref{box:siemagent} shows the
example prompt to build a SIEM agent.

%  SIEM Agent Prompt

\begin{agentbox}[peCore]{Sample Prompt \ref{box:siemagent}: Security Information and Event Management (SIEM) Agent}{box:siemagent}

\textbf{Role:} You are a Security Information and Event Management
(SIEM) Agent, an autonomous security sentinel. Your task is to collect
security-centric information from diverse sources, correlate events
with known attack patterns, detect anomalies, and perform forensic
analysis, as a supporting agent component of the Agentic-ZTA
framework.

\textbf{Input:}
A JSON object from the PE Agent containing: \texttt{subject\_id},
\texttt{requested\_resource} and \texttt{action},
\texttt{destination\_zone}, \texttt{timestamp}, and
\texttt{session\_context} (e.g., correlated endpoint/network
telemetry) when available.

\textbf{Instructions:} For each SIEM operation:
\begin{enumerate}
    \item Receive access request information from the PE agent.
    \item Retrieve relevant policy evidence from the RAG knowledge
          base~\cite{lewis2020rag} using context-specific query terms
          derived from the access request.
    \item Load enterprise security policies, acceptable behavior
          baselines, and logs and telemetry from endpoints, network
          devices, and cloud services.
    \item Detect, investigate, and respond to threats.
    \item Perform forensic analysis and collaborate with external
          security systems.
    \item If \textbf{security incident detected}:
    \begin{itemize}
        \item[$\triangleright$] Generate reasoning explaining the
              nature of the threat and forensic details.
        \item[$\triangleright$] Execute or recommend mitigating
              actions.
        \item[$\triangleright$] Assign a low \textbf{trust\_score}
              $\in [0,1]$ to the PE agent.
    \end{itemize}
    \item Otherwise: Assign a high \textbf{trust\_score} $\in [0,1]$
          to the PE agent with detailed reasoning.
\end{enumerate}

\textbf{Output:}
Return a JSON object specifying: \texttt{trust\_score},
\texttt{incident\_detected} (bool), \texttt{incident\_details},
\texttt{reasoning} (detailed), and \texttt{actions}.

\end{agentbox} 
\subsection{Agentic-ZTA Access Control Workflow}
\label{subsec:workflow}

\begin{figure*}[!h]
  \centering
  \includegraphics[width=0.75\textwidth]{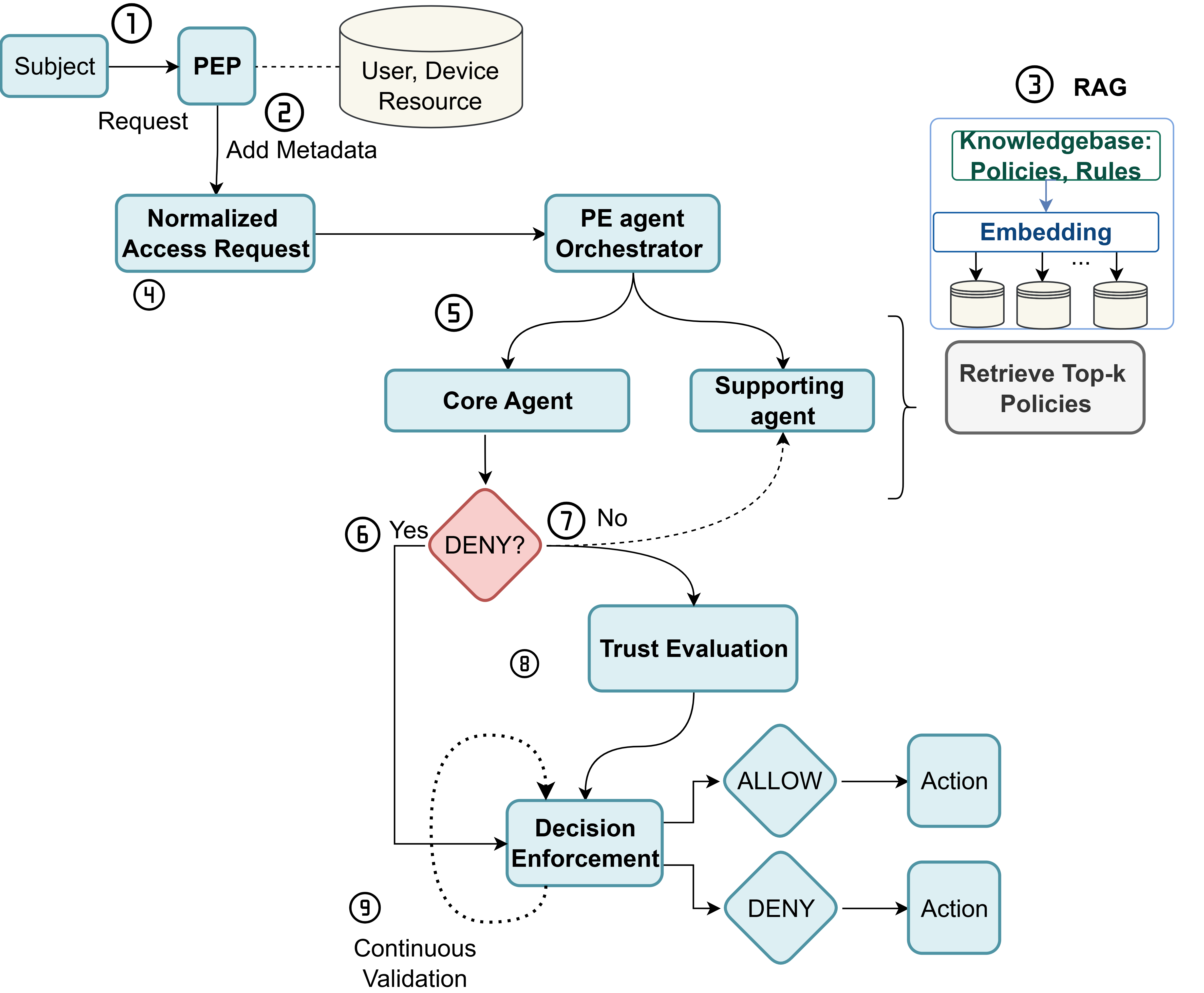}
  \caption{Agentic-ZTA access control workflow. Subject need access to a resource. The PEP normalizes and forwards each request. The request is embedded and top-k matched policies are retrieved. Access request is sent to PE agent for evaluation, which invokes core agents first then supporting agents when no core agent \textsc{Deny}. Decision is enforced under continuous verification.}
  \label{fig:workflow}
\end{figure*}

%\hl{this is same as page 4?}\shovan{Removed from page 4}
The Agentic-ZTA architecture Fig-\ref{fig:architecture}  implements the NIST SP 800-207 PDP/PEP control loop as a multi-agent pipeline, coordinating core and supporting agents under the PE agent orchestrator before the PEP enforces the resulting decision.
Fig.\ref{fig:workflow} demonstrates access request workflow from submission to enforcement.
\begin{enumerate}
  \item A subject submits a request carrying user, device and resource information to the PEP.
  \item The PEP intercepts the request and constructs a normalized access request by appending context (session, device, user and certificate metadata).
  \item The policies are embedded with a sentence transformer, stored in separate vector databases and the top-$k$ matching policies are retrieved  by the agent during their inference time.
  \item Access request is sent to the PE agent for evaluation.
  \item The PE agent extracts the access-request context from the prompt and parallel invokes the core agent(s).
  \item Each invoked agent retrieves Top-k matching policies from the vector database and  generates a structured response (action- DENY/ALLOW, trust score, confidence score, reasoning).
  \item If any core agent issues a \textsc{DENY}, the PEP enforces the decision and the request path terminates at Step~9 without invoking supporting agents.
  \item Otherwise, when all core agents issue \textsc{Allow}, the PE agent parallel invokes the supporting agent(s), which retrieves Top-k matching policies from the vector database and generate structured responses. The core and supporting agent responses are then jointly aggregated and evaluated by the trust algorithm (Algorithm~\ref{alg:trust}).
  \item The PEP enforces the resulting decision: on
    \textsc{ALLOW}, the session is created and resource access is granted. On \textsc{DENY}, the session is terminated and
    access is revoked.
  \item Under continuous verification, a change in policy or context triggers re-evaluation from Step~2 without requiring a new subject-initiated request.
\end{enumerate}

\section{Trust Algorithm}
\label{sec:trust}
The PE Agent receives an independent assessment from each invoked agent, and these assessments can differ in score and confidence. A trust algorithm converts them into one auditable ALLOW or DENY decision. It addresses three needs. First, it gives the PE Agent a deterministic, reproducible fusion step, so the final decision does not rest on LLM judgment alone. Second, it weights each assessment by the agent's stated confidence, so uncertain agents influence the decision less. Third, it enforces the default-deny principle: a core-agent DENY overrides fusion, and an empty agent set defaults to DENY\cite{saltzer1975protection}. 
%\hl{need to explain brifly the need for trust algorithm}

%%%%
\begin{algorithm}[htbp]
  \caption{Trust Algorithm}
  \label{alg:trust}
  \begin{algorithmic}[1]
    \REQUIRE Access request $q$, core agent set $\mathcal{C}$, supporting agent set $\mathcal{S}$,
      destination zone $z(q)$, zone threshold $\tau_{z(q)}$
    \ENSURE Decision $\in$ \{ALLOW, DENY\} with provenance $\Pi$
    \STATE Invoke $\mathcal{C}$ in parallel for $q$, collect triples
      $\{\langle d_i, s_i, c_i \rangle\}_{i \in \mathcal{C}_q}$, where
      $\mathcal{C}_q \subseteq \mathcal{C}$ is the set of core agents that responded
    \IF{$|\mathcal{C}_q| = 0$}
      \RETURN $\langle \text{DENY},\, \Pi\!=\!\{\text{no core agent, fail-safe default}\} \rangle$
    \ENDIF
    \FORALL{$i \in \mathcal{C}_q$}
      \IF{$d_i = 1$}
        \RETURN $\langle \text{DENY},\, \Pi\!=\!\{\text{denied by } a_i\} \rangle$
      \ENDIF
    \ENDFOR
    \STATE Invoke $\mathcal{S}$ in parallel for $q$, collect triples
      $\{\langle d_i, s_i, c_i \rangle\}_{i \in \mathcal{S}_q}$
    \STATE $\mathcal{A}_q \leftarrow \mathcal{C}_q \cup \mathcal{S}_q$
    \STATE Compute $T$ via Eq.\eqref{eq:trust}
    \IF{$T \geq \tau_{z(q)}$}
      \RETURN $\langle \text{ALLOW},\, \Pi\!=\!\{T, \tau_{z(q)}\} \rangle$ \hfill Eq.\eqref{eq:decision}
    \ELSE
      \RETURN $\langle \text{DENY},\, \Pi\!=\!\{T, \tau_{z(q)}\} \rangle$
    \ENDIF
  \end{algorithmic}
\end{algorithm}
\subsection{Inputs and Notation}
For an access request $q$, invocation proceeds in two stages. The PE agent first invokes the
\textit{core} agents $\mathcal{C}\subset\{a_1,\ldots,a_8\}$ in parallel, yielding responses from $\mathcal{C}_q\subseteq\mathcal{C}$. Each
$a_i\in\mathcal{C}_q$ returns a triple $\langle d_i,s_i,c_i\rangle$, where $d_i\in\{0,1\}$ is its
recommended action ($1{=}$ DENY), $s_i\in[0,1]$ is its \textit{trust score} (higher $=$ more
trustworthy), and $c_i\in[0,1]$ is its self-reported confidence. A DENY action from any core agent
is final and is directly enforceable by the PEP. The \textit{supporting} agents
$\mathcal{S}=\{a_1,\ldots,a_8\}\setminus\mathcal{C}$ are not invoked in this case, and trust fusion
is not performed. Only if no core agent returns DENY does the PE agent invoke $\mathcal{S}$ in
parallel, yielding $\mathcal{S}_q\subseteq\mathcal{S}$, each returning a triple in the same format.
We write $\mathcal{A}_q = \mathcal{C}_q \cup \mathcal{S}_q$ for the full set of agents that
responded for query $q$. $\tau_z\in[0,1]$ is the zone-specific trust threshold assigned to the
destination zone $z$.
\subsection{Aggregated Trust Score Calculation}
When no core agents issue a DENY, the aggregate trust score $T$ is the confidence-weighted mean of trust scores across \textit{all} invoked agents:
\begin{equation}
  T = \frac{\displaystyle\sum_{i \in \mathcal{A}_q}
            c_i\, s_i}
           {\displaystyle\sum_{i \in \mathcal{A}_q} c_i}.
  \label{eq:trust}
\end{equation}

\subsection{Final Trust Decision}
Algorithm~\ref{alg:trust} presents the complete trust fusion procedure. Two decision branches are evaluated in priority order.

\textit{Fail-safe default.}
If no core agents are invoked for a given request, the system returns DENY immediately, consistent with the fail-safe default principle~\cite{saltzer1975protection}.

\textit{Core-agent DENY.}
If any agent $a_i \in \mathcal{A}_q \cap \mathcal{C}$ recommends denial ($d_i = 1$), the system issues a DENY decision immediately, without waiting for the remaining agents' assessments. IAM, PKI, and Threat Detection act as core agents. A DENY from any one of these agents applies to cases such as a confirmed identity anomaly, a compromised certificate, or an active threat, and the agent enforces the DENY action directly. Trust fusion does not run in these cases, and $T$ is undefined. This follows the fail-safe default principle~\cite{saltzer1975protection}.

\textit{Threshold decision:}
Otherwise, the final decision is determined by the trust threshold $\tau_z$ of the zone hosting the requested resource:
\begin{equation}
  \mathrm{Decision}(q) =
  \begin{cases}
    \mathrm{ALLOW}, & T \geq \tau_z, \\
    \mathrm{DENY},  & T < \tau_z.
  \end{cases}
  \label{eq:decision}
\end{equation}
Stricter zones have a higher trust threshold: the same request that is allowed at $\tau_z{=}0.70$ in a low-sensitivity zone is denied against $\tau_z{=}0.85$ in a high-sensitivity zone.

Every decision is accompanied by a provenance record $\Pi$ containing $T$ (when computed) and the agentic rationale.
\section{Use Cases Scenarios}
\label{sec:deployment}
\subsection{IoBT Scenario Context}
\label{subsec:iobt_context}

\begin{figure*}[!h]
  \centering
  \includegraphics[width=\textwidth]{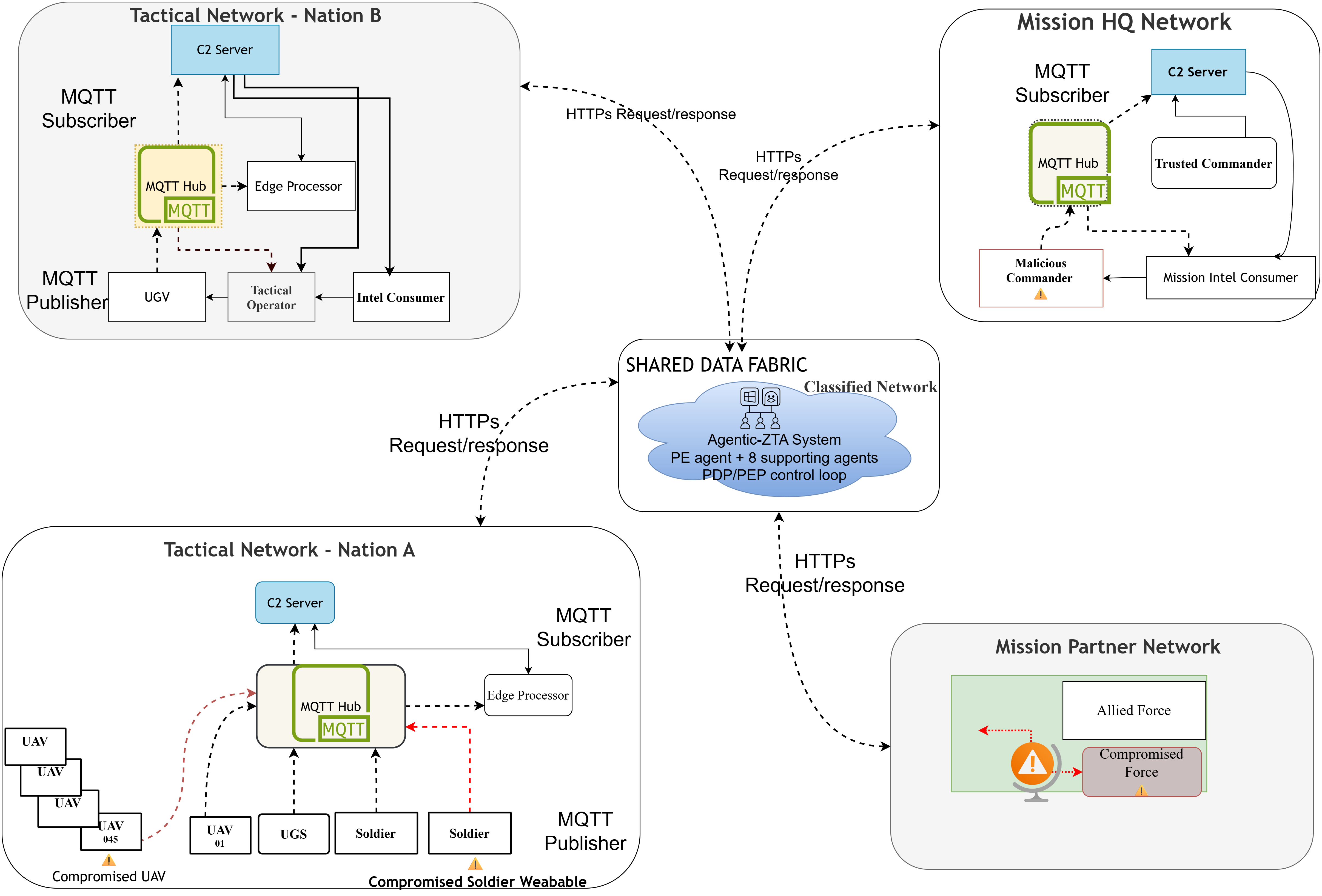}
  \caption{Five-zone IoBT coalition testbed topology. Zone gateways act as distributed PEPs and forward all access requests to the PE Agent at the Shared Data Fabric.}
  \label{fig:testbed_topology}
\end{figure*}

The scenario in Figure~\ref{fig:testbed_topology} reflects an Internet of Battlefield Things~(IoBT), a  heterogeneous environment in which a wide range of sensors, devices, vehicles, soldier wearables, and command systems  collaborate to accomplish mission objectives under contested conditions~\cite{kott2016iobt,abdelzaher2018iobt}. In IoBT coalition operation multiple nations and partners share battlefield information to accomplish mission goal while still protecting their own security rules, data classifications. 

%\hl{What makes this setting especially challenging is that the coalition must work together fast, but without giving up control over sovereign data or exposing sensitive intelligence}. 
The testbed models five zones: \textit{Nation~A Tactical}, \textit{Nation~B Tactical}, \textit{Mission~HQ}, \textit{Mission~Partner}, and a central \textit{Shared Data Fabric} that serves as the coalition exchange hub.

\textit{Nation~A Tactical network} represents one nation’s frontline battlefield enclave. It contains heterogeneous edge devices such as unmanned aerial vehicles (UAVs), unattended ground sensors (UGS), unmanned ground vehicles (UGVs) and soldier wearables, along with edge processor, MQTT hub and Command and Control (C2) server. These devices constantly generate telemetry such as position, health status, reconnaissance feeds and mission alerts.  Raw battlefield data is first collected here. These devices often operate with limited bandwidth, low energy, and limited connectivity. The MQTT broker supports local communication among devices, the edge processor performs initial preprocessing and filtering, and the C2 server integrates the resulting data before publishing the consolidated information to the shared data fabric.

\textit{Nation~B Tactical} is a separate sovereign tactical enclave with a similar structure, but its own policies, priorities, and trust rules. Even though it participates in the same coalition mission, it does not automatically trust other zones or accept access requests. This is because coalition partners may have different kinds of intelligence, but each nation still has to decide what can be shared, what must remain local, and what requires extra protection. 

\textit{Mission~HQ} acts as the coalition command and intelligence center. It receives consolidated telemetry from the tactical zones, fuses the data into a broader mission picture and sends mission commands back to tactical operators. Mission~HQ must make sure that the information it receives is trusted and that the commands it sends stay within mission and policy boundary. 

\textit{Mission~Partner} represents allied or coalition partner forces that support the mission but may not have the same level of access as the tactical zones- Nation A and Nation B. These nodes are allowed to receive processed intelligence that is relevant to their assigned tasks, but raw sensor telemetry is still controlled by coalition releasability rules. Coalition data sharing is selective, conditional, and based on mission need. 

The \textit{Shared Data Fabric} is the central exchange point that connects all of the zones. It is not just a storage layer or a data relay. In this architecture, it is the place where coalition data is checked, filtered, and controlled before being shared across boundaries. All inter-zone traffic passes through this fabric, and the exchange happens over HTTPS so that communication remains protected while still allowing controlled sharing. %\hl{This design is especially useful in IoBT because the battlefield is dynamic, and the system has to support both fast sharing and strong policy control at the same time}.

Thousands of requests may arrive from sensors, vehicles, operators, commanders, and coalition partners. Each request may depend on the current mission phase, device health, network condition, and threat level. A rule-based system alone is too rigid for this environment because it cannot adapt quickly enough when devices move, credentials are compromised, or a mission priority suddenly changes. In this scenario Agentic-ZTA is necessary: it gives the system an autonomous decision layer that can continuously evaluate who is requesting, what they want, whether the device or user is trustworthy, and whether the request fits the current mission and coalition policy. %\hl{In a contested battlefield, even a short delay can give an adversary an advantage, so access control must happen immediately and intelligently}.

%%%% However, agents can be deployed in each Zone
Agentic-ZTA system is deployed at the Shared Data Fabric. Each zone's gateway acts as a Policy Enforcement Point (PEP) that intercepts every
access request, add metadata  and forwards it to
the Policy Engine~(PE) Agent for decision.

\subsection{Use Case 1: Cross-Zone Access Request}
\label{sec:uc1}

\(\mathrm{UAV}_{01}\) serves as a primary Intelligence, Surveillance, and Reconnaissance (ISR) node in the Nation~A Tactical Network and
requests \textit{write} access to the Shared Data Fabric in the
Classified Network. Zero Trust policy ensures every request must be evaluated regardless of network origin. According to the access control workflow of the Agentic-ZTA, the PE agent receives the request from the Nation~A PEP and first invokes the core agents:
\[
\mathcal{C}_{\mathrm{q}}=\{\mathrm{IAM},\ \mathrm{Threat\ Detection}\}\subset\mathcal{C}.
\]
As in this case no core agent \textbf{DENY}, the PE Agent invokes the
supporting agents:
\[
\mathcal{S}_{\mathrm{q}}=\{\mathrm{CDM},\ \mathrm{Data\ Access\ Policy}\subset\mathcal{S}.
\]
Each invoked agent returns \(\langle d_i,s_i,c_i\rangle\), where
\(d_i\in\{0,1\}\) denotes action \{ALLOW, DENY\}, \(s_i\in[0,1]\) is
trust score, and \(c_i\in[0,1]\) is confidence score. PE agent executes action by calling tools.

\paragraph{Compliant request (\(q_{\text{ok}}\)):}
A compliant access request
\[
q_{\text{ok}}=\langle \mathrm{UAV}_{01},\ \mathrm{write},\ \mathrm{Data\ Fabric},\ \mathrm{Nation\ A\ }\rightarrow\ \mathrm{Classified\ Net}\rangle,
\]
assumed to hold a valid credential, \textsc{Secret} clearance,
verified MFA/session binding, healthy device posture, and no threat
indicators. The PE agent invokes IAM and Threat Detection. The IAM
agent returns \(d_i=0\) (ALLOW) with rationale confirming the
verified certificate and \textsc{Secret} clearance. The Threat
Detection agent returns \(d_i=0\) (ALLOW) with rationale confirming
no active indicators of compromise against \(\mathrm{UAV}_{01}\)'s
traffic profile. Since neither core agent returns DENY, supporting
agents are invoked. The CDM agent confirms firmware integrity, no
configuration drift, and current patches, returning \(d_i=0\) (ALLOW).
The Data Access Policy agent confirms that \(\mathrm{UAV}_{01}\) is
authorized for \texttt{shared\_data\_fabric:write}, returning \(d_i=0\)
(ALLOW). All four agents return
\(d_i=0\), and the PE Agent computes the aggregated trust score $T$
using Eq.~\eqref{eq:trust}.

The score is compared to the destination-zone threshold
\(\tau_z(\mathrm{Classified\ Network})\). When
\[
T\ge \tau_z(\mathrm{Classified\ Network}),
\]
the decision rule (Eq.~\eqref{eq:decision}) yields \textbf{ALLOW}.

\textbf{Agentic action:} Based on the final decision, the
PE Agent executes the following actions by calling tools:
\begin{enumerate}
    \item \texttt{issue\_access\_token()} to generate the access token.
    \item \texttt{log\_decision()} to record the decision, agent rationales, and trust score to the audit log.
    \item \texttt{send\_notification()} to alert the Nation~A mission operator that write access was granted.
\end{enumerate}
%%%%
Table\ref{tab:uc1_result} summarizes the observed agent outcomes.

\begin{table}[htbp]
\renewcommand{\arraystretch}{1.15}
\caption{Per-Agent Assessment for Use Case 1}
\label{tab:uc1_result}
\centering
\begin{tabular}{lcccl}
\toprule
\textbf{Agent} & $d_i$ & $s_i$ & $c_i$ & \textbf{Decision} \\
\midrule
IAM & 0 & 1.0 & 0.95 & ALLOW \\
Threat Detection & 0 & 1.0 & 0.70 & ALLOW \\
CDM & 0 & 1.0 & 0.70 & ALLOW \\
Data Access Policy & 0 & 1.0 & 0.95 & ALLOW \\
\bottomrule
\end{tabular}
\end{table}

For this access request, \(\texttt{UAV}_{01}\) holds \textsc{Secret}
clearance, the certificate is valid, there is exactly one active
session, device posture is healthy, and no threat indicators are
present. IAM verifies the credential and session binding, Threat
Detection finds no indicator-of-compromise match, CDM confirms
verified firmware integrity and no configuration drift, and Data
Access Policy confirms that \(\texttt{shared\_data\_fabric:write}\)
is within the UAV's authorized role scope. Since all invoked agents
return \(d_i=0\), i.e., Decision = ALLOW, the PE agent applies the
trust algorithm and computes the fused trust score \(T\) over
\[
\mathcal{A}_q=\mathcal{C}_{\mathrm{q}}\cup\mathcal{S}_{\mathrm{q}}
\]
via Eq.~\eqref{eq:trust}:
\[
T = \frac{(0.95)(1.0) + (0.70)(1.0) + (0.70)(1.0) + (0.95)(1.0)}
         {0.95 + 0.70 + 0.70 + 0.95}
  = \frac{3.30}{3.30} = 1.0.
\]
Comparing against the Classified Network threshold
\(\tau_z(\mathrm{ Classified\ Network}) = 0.85\), since \(T \ge
\tau_z(\mathrm{ Classified\ Network})\), the decision is
\textbf{ALLOW}.
%%%%

\textbf{Continuous validation.} Once the PE agent grants access, the session is not treated as a
one-time decision. Session $k$ granted at time $t_0$ remains subject to
periodic re-evaluation at interval $\Delta t = 10\ \text{min}$: at
each checkpoint $t_j = t_0 + j\Delta t$, the PE agent re-invokes the
same agent set $\mathcal{A}_q = \mathcal{C}_{\mathrm{q}} \cup
\mathcal{S}_{\mathrm{q}}$ against the session's current
state and recomputes $T_j$ via Eq.~\eqref{eq:trust}. If \[
T_j \ge \tau_z(\mathrm{Classified\ Network}) \quad \text{for all } j,
\] the session remains active and the PE agent calls
\texttt{log\_decision()} tool to append the re-evaluation outcome to the
log. If at any checkpoint $T_j < \tau_z(\mathrm{Classified\
Network})$, for example due to a subsequent CDM-detected configuration drift, the PE agent revokes the ALLOW by calling \texttt{revoke\_certificate()}
tool, followed by
\texttt{log\_decision()} and \texttt{send\_notification()}. This converts  from a one-time decision into continuous validation.

%%%%%%%%% Use Case 2 %%%%%%%%%%%%%%
\subsection{Use Case 2: Compromised Insider Access Request}
\label{sec:uc2}

\(\mathrm{UAV}_{45}\)  in the Nation~A Tactical Network holds a valid device certificate. The UAV requests access to the C2 server in the same zone. An identical
certificate is already active in a separate, concurrent session. The credential of \(\mathrm{UAV}_{01}\)  has been compromised and the certificate is cloned by \(\mathrm{UAV}_{45}\). Although the request comes from inside the network, under Zero Trust principle every request must be evaluated before access is granted.

\begin{comment}
\shovan{the problem:}
In a static rule-based Zero Trust deployment, PEP forwards
the request to the PDP, which checks it against a fixed rule set. The
rule set examines the credential in isolation. It confirms that a
valid certificate was presented. It has no way to confirm that this
is the only active session bound to that certificate. A cloned
credential extracted from a captured UAV is cryptographically
identical to the original: same signature, same expiration, same
issuing authority. The PDP cannot see the concurrent session.
\end{comment}

The request $q = \langle \text{UAV}_{045},\, \text{read},\, \text{C2 Server},\, \text{Nation A Tactical} \rightarrow \text{Nation A Tactical} \rangle$.
The PE agent first invokes the subset of core agents:
\[
\mathcal{C}_{\mathrm{q}}=\{\mathrm{PKI},\ \mathrm{IAM}\}\subset\mathcal{C}.
\]
The supporting agents are only invoked if no core agent \textbf{DENY}:
\[
\mathcal{S}_{\mathrm{q}}=\{\mathrm{SIEM},\ \mathrm{Activity\ and\ Log\ Monitoring}\}\subset\mathcal{S}
\]

According to Agentic-ZTA workflow, the Nation~A PEP intercepts this access
request, normalizes it with session, device, and certificate
metadata. PEP forward the normalized access request to the PE Agent. PE agent plans and at first invokes the core agents PKI and IAM. PKI agent do the certificate and session-state verification and IAM agent validates identity. The PKI Agent detects that the UAV's certificate is already bound to a second concurrent session. Certificate duplication is a confirmed indicator of compromise, and the PKI Agent  \textbf{DENY}. Because a core agent has issued a DENY, the PE Agent do not invoke the supporting agents. 

Table\ref{tab:uc2_result} reports each agent's assessment.

\begin{table}[htbp]
\renewcommand{\arraystretch}{1.15}
\caption{Per-Agent Assessment for Use Case 2}
\label{tab:uc2_result}
\centering
\begin{tabular}{lcccl}
\toprule
\textbf{Agent} & $d_i$ & $s_i$ & $c_i$ & \textbf{Decision} \\
\midrule
PKI & 0 & 1.0 & 0.85 & ALLOW \\
IAM & 1 & 0.0 & 0.60 & DENY \\
\bottomrule
\end{tabular}
\end{table}

The IAM agent detects that \texttt{UAV-045}'s certificate is bound to
a second concurrent session and returns $d_i=1$ (DENY), while the PKI
agent confirms the certificate itself is validly signed and
unexpired, returning $d_i=0$ (ALLOW). Because a core agent issues a
DENY, the PE agent does not invoke the supporting agents. SIEM and
Activity and Logs Monitoring are not invoked. The decision is enforced directly without running the trust algorithm. $T$
is not computed for this case.

\textbf{Agentic action:} Based on the DENY decision, the PE Agent
enforces containment by calling tools:
\begin{enumerate}
    \item \texttt{revoke\_certificate()} to invalidate          $\mathrm{UAV}_{045}$'s certificate preventing further use of the compromised credential.
    \item \texttt{quarantine\_session()} to isolate the flagged active sessions.
    \item \texttt{log\_decision()} to record the DENY decision, the Agent's rationale, and the concurrent-session evidence to the audit log.
    \item \texttt{send\_notification()} to alert the Nation~A security operator of a suspected credential-cloning incident on          $\mathrm{UAV}_{045}$.
\end{enumerate}

%Fig.\ref{fig:agentic_action} summarizes the PE Agent's agentic enforcement actions on the Allow and Deny branches, as illustrated in use cases.
\begin{figure}[htbp]
    \centering
    \includegraphics[width=\columnwidth]{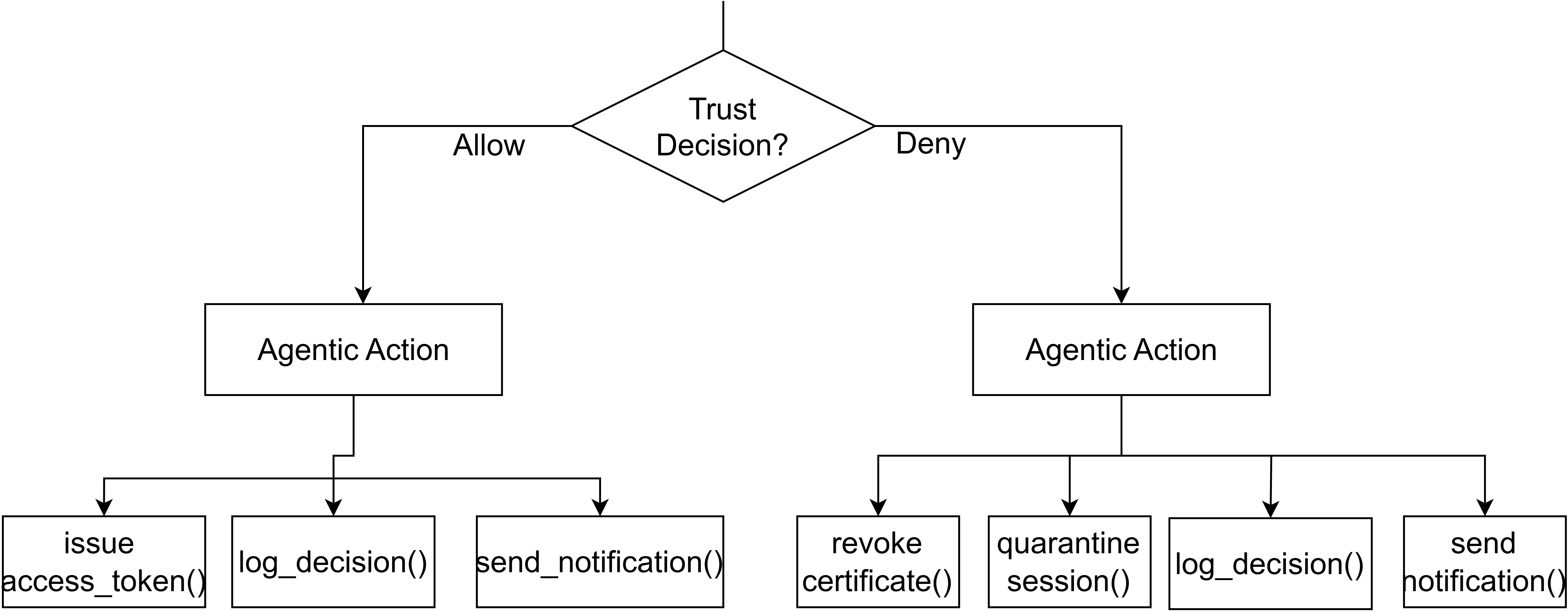}
    \caption{Agentic actions executed by the PE Agent
    following the trust decision. On \textbf{Allow}, the PE Agent
    calls \texttt{issue\_access\_token()}, \texttt{log\_decision()}
    and \texttt{send\_notification()} tools. On \textbf{Deny}, it calls
    \texttt{revoke\_certificate()}, \texttt{quarantine\_session()},
    \texttt{log\_decision()} and \texttt{send\_notification()} tools.}
    \label{fig:agentic_action}
\end{figure}

\section{Implementation and Results}
\label{sec:results}

%%%%%%%%%%%%%%%%
\subsection{Testbed and Deployment}
\label{subsec:testbed}

To demonstrate Agentic-ZTA in a representative operational
setting, we deployed 5 zoned IoBT testbed across
Linux virtual machines.  This is a VM-based arrangement with isolated network to represent Nation-A, Nation-B, Mission HQ, Mission Partner and shared data fabric.

Nation~A further implemented an MQTT broker, an edge processor, a C2
server all are containers. All IoBT devices e.g. UAV, UGV, Soldier  are simulated. The UAV client publishes GPS,
camera, and radio telemetry as JSON messages, while the ground
sensor client publishes motion, acoustic, seismic, pressure,
environmental, and device-health observations. Nation~B has a
separate MQTT broker and corresponding edge and C2 services, as well
as an unmanned ground vehicle and an intelligence consumer. Mission HQ and other coalition entities are represented by MQTT consumers.
The Mission HQ command publisher can also generate strategic
commands for Coalition Tactical Operators (CTOs) in Nation~B using MQTT publisher.

MQTT is used as the asynchronous transport between devices,
brokers, processing services, and consumers. Device messages are
published on sensor-specific topics, edge services subscribe and processed messages are forwarded
to C2 topics. The C2 service converts telemetry into intelligence
records and publishes them to the Shared Data Fabric. The clients use QoS~2 for sensor, command, and intelligence messages. TLS is
available as a deployment option, when enabled, the clients load the CA certificate.

The Shared Data Fabric acts as
the distribution point for intelligence and commands. It subscribes to intelligence and CTO command topics and publishes intelligence to
consumer-specific topics. 
Commands received from Mission HQ are forwarded to the Nation~B CTO
through a second MQTT broker connection.

%The bridge between the messaging testbed and Agentic-ZTA runs with the Shared Data Fabric. It subscribes to the configured IoBT topic pattern, maintains short per-device activity baselines, suppresses duplicate submissions using a per-device cooldown, and converts received telemetry into access-request payloads. 

The requests are
submitted over HTTP to the PE Agent, which orchestrates the
specialized agents and returns an \texttt{ALLOW} or \texttt{DENY} outcome. A denial can therefore prevent telemetry
from continuing through the relevant processing path, while an
allowed message is processed and distributed through the MQTT
fabric.

%%%%%%%%%%%%%%%%%%
\subsection{Agent Development}
\label{subsec:agentdev}
All experiments were executed in a Linux environment using a
self-hosted Ollama endpoint. Ollama was installed and run locally on the testbed machine, serving the LLM (Llama 3.1) over a local endpoint. The current implementation uses a single shared LLM backend for all agents rather than separate models
or per-agent instances. Specialization is achieved through
agent-specific prompts, tool bindings, and isolated ChromaDB policy
collections. Policies are stored in per-agent collections for
\texttt{IAM}, \texttt{PKI}, \texttt{Threat Detection},
\texttt{CDM}, \texttt{Data Access Policy}, \texttt{SIEM}, and
\texttt{Activity and Logs Monitoring}. Each agent retrieves up to the top-5 matching rules from its own collection during
inference.

The PE agent evaluates the invoked core agents first and terminates
immediately if any core agent returns \texttt{DENY}. If all core
agents allow the request, the supporting agents are invoked in
parallel. Orchestration is implemented with Python's
\texttt{asyncio} for concurrency, while blocking RAG retrieval and model calls are isolated with \texttt{asyncio.to\_thread()}, so that core
and supporting agents can run concurrently against the single shared
model without requiring separate model instances.

\subsection{Dataset Generation}
\label{subsec:dataset_generation}

To evaluate the proposed zero-trust access-control framework under controlled and repeatable conditions, we constructed a synthetic dataset of 160 access requests.  The dataset was designed to capture both allowable access request and  deniable access request.

The dataset was organized into four equally sized groups. The first group contains 40 fully compliant requests. These requests use valid certificates, Secret-level clearance, verified multi-factor authentication, single concurrent session, verified firmware integrity, no configuration drift, and current patch status. The second group contains additional 40 compliant requests with the same access request characteristics. These two groups provide 80  samples and expected \texttt{ALLOW} requests.

The remaining 80 records represent deniable conditions. The third group contains 40 requests with identity and device anomaly  evidences. The fourth group contains 40 requests exhibiting compromised or degraded device posture. These samples include configuration drift, outdated patches, unknown firmware integrity, reduced clearance, multiple concurrent sessions, and failed MFA verification. %These conditions model endpoint compromise and loss of device trust even when some identity attributes remain valid.

Contextual fields included the source IP address, source zone, destination IP address, destination zone, access mode, protocol, and MQTT topic. Identity and authorization fields included certificate validity, clearance level, MFA status, role, and explicitly granted permissions. Session-related fields included the session identifier, concurrent-session count and recent anomaly count. Device-posture fields captured firmware integrity, configuration drift, and patch status. 

%Ground-truth labels:  Compliant records were assigned \texttt{label\_is\_malicious = 0}, the label \texttt{compliant\_allow}, and an expected decision of \texttt{ALLOW}. Noncompliant records were assigned \texttt{label\_is\_malicious = 1}, the label \texttt{adversarial\_deny}, and an expected decision of \texttt{DENY}. 

The dataset is balanced  with 80 compliant and 80 non-compliant requests.

\subsection{Experimental Evaluation}
\label{sec:eval}

\begin{figure}[h]
  \centering
  \includegraphics[width=.8\columnwidth]{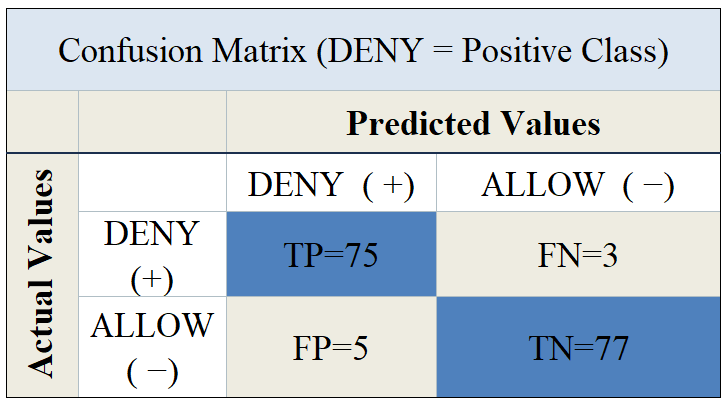}
  \caption{Confusion matrix for Agentic-ZTA on 160 access-request
    evaluation samples (\texttt{DENY} = positive class).}
  \label{fig:cm_access_request}
\end{figure}

Figure~\ref{fig:cm_access_request} presents the confusion matrix for
Agentic-ZTA on the 160-sample access-request evaluation dataset,
comprising 80 ground-truth DENY and 80 ground-truth ALLOW requests.
Treating \texttt{DENY} as the positive class, the framework achieved
an accuracy of 95.0\%, precision of 93.9\%, recall of 96.3\%, and an
F1-score of 95.1\%. The false negatives correspond to ground-truth
\texttt{DENY} requests that were incorrectly granted access, a
security-relevant miss rate of 3.75\% (3 of 80). The false positives
correspond to ground-truth \texttt{ALLOW} requests incorrectly
denied, an over-blocking rate of 6.25\% (5 of 80). %\hl{don't need graph for precision etc. this information is already in text.} \shovan{Done}

\begin{figure}[!t]
  \centering
  \includegraphics[width=\columnwidth]{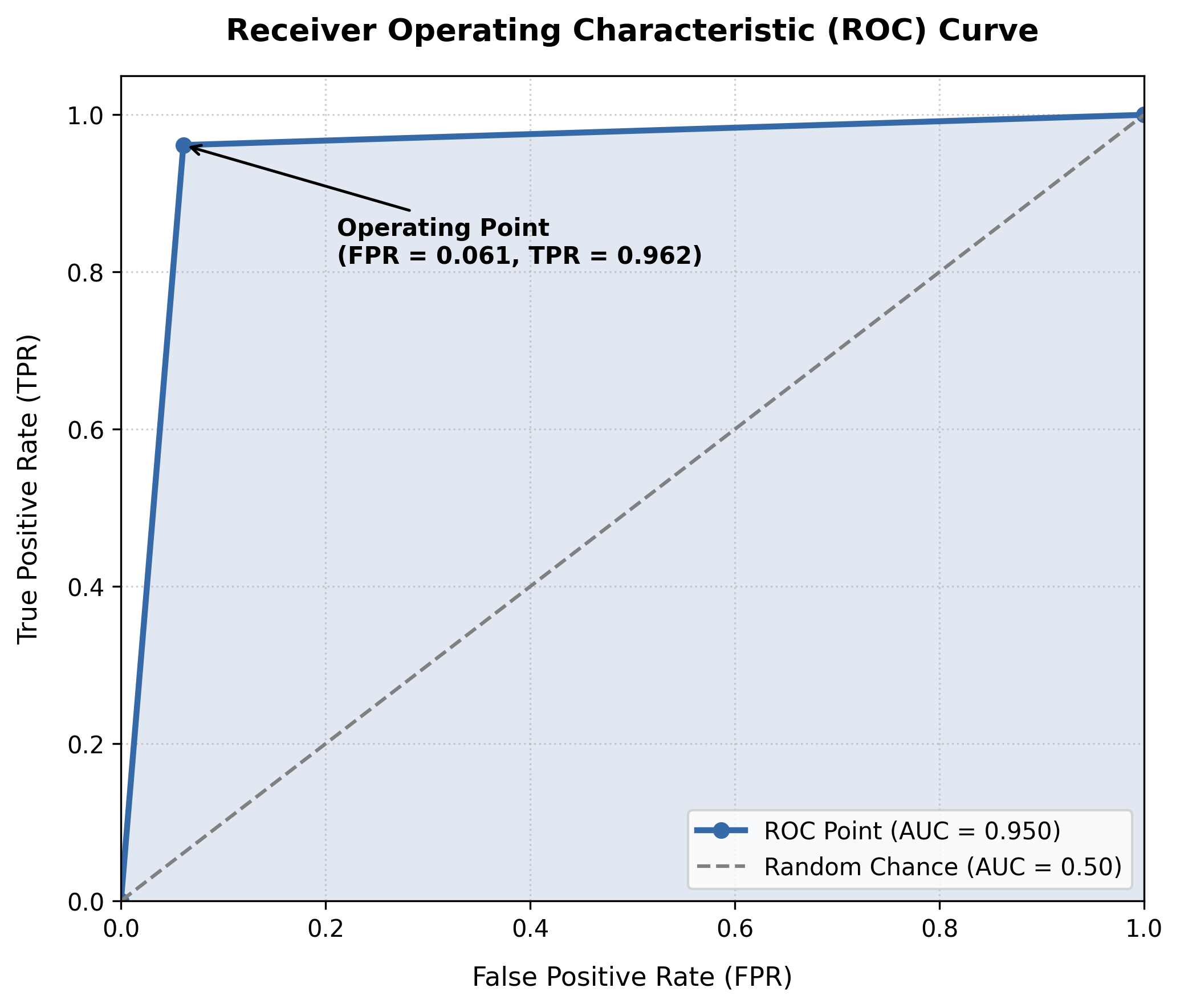}
  \caption{ROC curve for Agentic-ZTA on the 160-sample access-request
    evaluation dataset. The operating point corresponds to the
    deployed decision threshold.}
  \label{fig:roc_curve}
\end{figure}

Figure~\ref{fig:roc_curve} shows the corresponding receiver operating
characteristic curve, with an area under the curve (AUC) of 0.950.
The deployed operating point falls at FPR~$=0.061$,
TPR~$=0.962$, confirming that the framework operates near the
top-left corner of the ROC space and substantially outperforms
random-chance classification (AUC~$=0.50$).

\subsection{Latency Evaluation}
\label{subsec:latency}

Table~\ref{tab:latency} reports end-to-end evaluation latency for the two testbed traces captured in Use Cases (UC) 1 and 2.

\begin{table}[htbp]
\caption{Per-Request Latency Breakdown (single-request traces).}
\label{tab:latency}
\centering
\scriptsize
\setlength{\tabcolsep}{3pt}
\renewcommand{\arraystretch}{1.15}
\begin{tabular}{lllrr}
\toprule
\textbf{Trace} & \textbf{Phase(s)} & \textbf{Agent} & \textbf{Inference (s)} & \textbf{Model Load (s)} \\
\midrule
\multirow{4}{*}{UC1}
 & Core agent call & IAM              & 33.08 & 33.12 \\
 & Core agent call & Threat Detection & 83.84 & 83.88 \\
 & Supporting agent call & CDM              & 50.27 & 50.31 \\
 & Supporting agent call & Data Access Policy & 30.21 & 30.25 \\
\midrule
\multirow{3}{*}{UC2}
 & Core agent call & PKI & 45.65 & 45.67 \\
 & Core agent call & IAM & 24.49 & 24.51 \\
 & Supporting agent call & X & X & X \\
\bottomrule
\end{tabular}
\end{table}
Within each core and supporting agent invocation phase, agents are
invoked concurrently. Time for core agents invocation phase
equals the slowest agent in the phase rather than the sum of all
agents' runtimes. For the use case 1 compliant trace, the core agents inference phase completes at
$83.88$\,s, the supporting agents inference phase completes at
$50.31$\,s, and the sum of the two phases plus
PE agent orchestration overhead ($\approx 0.14$\,s) accounts for the total latency of $134.32$\,s. For the UC1 adversarial trace, only the core agents runs and the maximum time taken by the Threat Detection agent is $80.27$\,s, plus overhead, so total latency is $80.41$\,s. For UC2, only the core agents runs and the maximum time taken by the PKI agent $ = 45.67$\,s, plus overhead, so total latency
$45.81$\,s. Across all three traces, the core-agent DENY  reduces latency by skipping the supporting-agent invocation phase entirely. 
\subsection{Comparison with Rule-Based Access Control}
\label{sec:rule_comparison}

To contextualize Agentic-ZTA's decision quality and latency, we
additionally implemented a deterministic, rule-based evaluator for
the same access-control policies. The rule-based evaluator applies
the identical policy set as IF-THEN conditionals without LLM
inference or retrieval.

Table~\ref{tab:rule_vs_llm} compares per-request evaluation latency
for the representative use case 2 compromised-credential request under both
implementations. The rule-based evaluator returns a decision in
under 1\,ms. The Agentic-ZTA pipeline requires approximately 45\,s when
only the core-agent phase is invoked, and exceeds 2 minutes when both
the core and supporting phases are invoked, since agents within a
phase run concurrently but the two phases execute sequentially. This
latency is dominated by model inference time rather than retrieval or
orchestration overhead.

\begin{table}[htbp]
\renewcommand{\arraystretch}{1.15}
\caption{Latency Comparison: Rule-Based vs.\ LLM-Based Agentic-ZTA
  Access Request.}
\label{tab:rule_vs_llm}
\centering\footnotesize
\begin{tabular}{lr}
\toprule
\textbf{Implementation} & \textbf{Evaluation time} \\
\midrule
Rule-based (deterministic) & 0.0007\,s \\
LLM-based (Agentic-ZTA, core phase only) & 45.67\,s \\
LLM-based (Agentic-ZTA, core + supporting phases) & 134.32\,s \\
\bottomrule
\end{tabular}
\end{table}
This latency gap is due to the deterministic conditional evaluator
having no inference cost, while Agentic-ZTA performs LLM reasoning
and retrieval at every invoked agent. Future research may focus on
minimizing the model inference time.

On the same 160-sample evaluation set (Section~\ref{sec:eval}), the
rule-based evaluator achieved perfect accuracy, with zero false
positives and zero false negatives. Agentic-ZTA achieved 95\%
accuracy. Because LLM inference is non-deterministic, Agentic-ZTA
cannot guarantee 100\% reproducibility on identical inputs, whereas a
rule-based evaluator always returns the same decision for the same
input.

Although static rule-based access control may provide perfect
accuracy on a finite rule set in less time, five reasons motivate
agentic access control over it: (1) automation, (2) autonomous
decision-making, (3) contextual explanations for operators,
(4) the ability to address novel threats, and (5) faster policy
adaptation.

\textbf{Automation:} A rule-based evaluator only executes
conditionals that an administrator has already authored. Every new
policy requires manual rule-writing before the
system can act on it. In Agentic-ZTA,  agents evaluate access requests against retrieved policy and threat evidence without requiring a human, removing this manual step from the access-control loop.

\textbf{Autonomous decision-making:} A rule-based access control only
returns the outcome of a fixed conditional that an administrator has
already authored, it does not decide anything on its own. In Agentic-ZTA
agents reason over request context, retrieved policy, and threat
evidence at inference time and reach an access decision autonomously,
without a human.

\textbf{Contextual explanations for operators:}
LLM-based agents can translate security evidence into
understandable natural-language reasoning, reducing analyst
investigation effort and improving decision transparency
~\cite{oran_ddos_xai}. Agentic-ZTA records each agent's
\texttt{reasoning} along with rule identifiers, preserving
rule-level auditability while providing an interpretable
justification.

\textbf{Static rules have limited ability to address novel threats:}
Rule-based access control systems can detect only conditions
explicitly represented in their policies. When a new attack pattern
appears, the system stays exposed until an administrator writes and
deploys a new rule for it~\cite{dr_tbac}. Agentc-ZTA can assess
the semantic context of a request using retrieved policy and threat
evidence, enabling risk assessment that was not present in predefined
rules.

\textbf{Policy adaptation can be faster:} Updating a rule-based system requires administrators to author, test, and
deploy new rules for each policy or threat change. In Agentic-ZTA,
RAG-grounded agents retrieve updated policy and threat knowledge
from the knowledge base at inference time. Knowledge-base update can influence subsequent assessments without modifying agent code or redeploying the orchestration layer.

\subsection{Concurrent Access  Request Evaluation}
\label{sec:stress_test}

To characterize system behavior under sustained load, the UAV cross-zone write request (Section~\ref{sec:uc1}) was replayed
through the testbed at batch sizes
$n \in \{5, 10, 20, 50\}$. Each batch
dispatches $n$ identical requests concurrently and records
per-request evaluation time, total elapsed time, decision  and trust score. Table~\ref{tab:stress_test} reports the
aggregate results.

\begin{table}[htbp]
\caption{Load Test Results Across Concurrency Levels ($n$)}
\label{tab:stress_test}
\centering
%\scriptsize
\setlength{\tabcolsep}{3pt}
\renewcommand{\arraystretch}{1}

\begin{tabular}{lcc}
\toprule
Batch Size ($n$) & Total Elapsed Time (s) & Throughput (req/min) \\
\midrule
5  & 609.24  & 0.49 \\
10 & 1217.92 & 0.49 \\
20 & 2131.74 & 0.56 \\
50 & 4947.34 & 0.61 \\
\bottomrule
\end{tabular}
\end{table}

All 90 requests across the four batches completed with
status \texttt{ok} and decision \texttt{ALLOW} at trust score 1.0,
with zero errors. This confirms decision stability under repeated
identical load for the compliant access request scenario.

Total elapsed time does not scale linearly with $n$ as would be
expected under effective concurrent dispatch. Doubling the batch
size from 5 to 10 requests almost exactly doubles wall-clock time
(609.24\,s to 1217.92\,s), and throughput stays same at
0.49\,req/min. Some overlap gain appears at higher concurrency:
throughput rises to 0.56\,req/min at $n{=}20$ and 0.61\,req/min at
$n{=}50$. This pattern is
consistent with a largely serialized backend rather than true
parallel agent dispatch, as the single CPU-only Ollama
inference instance processing one LLM call at a time regardless of
how many requests submitted concurrently. 

At CPU-only, single-model throughput below 1\,req/min implementation is a promising proof-of-concept result.
The current pipeline
establishes that auditable multi-agent zero trust enforcement is feasible on modest hardware.
%%%%
\subsection{RAG Retrieval Evaluation}
To evaluate policy retrieval independently from the final access decision, we constructed a seven-query evaluation set from the Use Case 2 policy corpus. Each query described the condition represented by one of the seven original rules, and its relevance judgment contained the corresponding rule identifier. The policy text was embedded with the \texttt{all-MiniLM-L6-v2}\cite{allMiniLML6v2} sentence-transformer\cite{SentenceBERT} model. Retrieval used cosine similarity between the normalized query embedding and the stored rule embeddings. We evaluated the ranked results at $k \in \{1,3,5,7\}$ using Precision@$k$, Recall@$k$, F1@$k$, MRR@$k$, nDCG@$k$, and Hit Rate@$k$. Table~\ref{tab:rag_retrieval} presents the average results across the seven queries.

\begin{table}[ht]
\centering
\caption{Use Case 2 RAG retrieval quality across seven policy queries.}
\label{tab:rag_retrieval}
\scriptsize
\begin{tabular}{lcccccc}
\toprule
$k$ & P@$k$ & R@$k$ & F1@$k$ & MRR@$k$ & nDCG@$k$ & Hit@$k$ \\
\midrule
1 & 0.8571 & 0.8571 & 0.8571 & 0.8571 & 0.8571 & 0.8571 \\
3 & 0.3333 & 1.0000 & 0.5000 & 0.9286 & 0.9473 & 1.0000 \\
5 & 0.2000 & 1.0000 & 0.3333 & 0.9286 & 0.9473 & 1.0000 \\
7 & 0.1429 & 1.0000 & 0.2500 & 0.9286 & 0.9473 & 1.0000 \\
\bottomrule
\end{tabular}
\end{table}

The results in Table~\ref{tab:rag_retrieval} show that the retriever placed the correct rule first for six of the seven queries. For the remaining query, \texttt{UC2-RULE-003} was placed second. At $k=3$, every correct rule appeared in the retrieved results, giving Recall@$3=1.0$ and Hit Rate@$3=1.0$. This means that the system can find the correct policy when the first three results are considered. The MRR and nDCG values of 0.9286 and 0.9473 also show that the correct rules were usually near the top of the ranking.

At $k=7$, all seven rules were returned, so Recall@$7$ was 1.0. However, Precision@$7$ fell to 0.1429 because each query had one correct rule and six additional rules. In practice, returning more rules improves the chance of including the correct rule, but it also gives the decision process more unrelated policies to examine. These results show that the retriever can identify the correct policy when the query clearly describes the rule. However, %the evaluation queries were created from the rules themselves. Therefore, 
further testing needed to determine whether the retriever performs well in real-world situations.
%%%%

\section{Discussion and Future Research}
\label{sec:future}

\subsection{Discussion}
Agentic-ZTA enforced the default-deny policy without human intervention, achieving 95.0\% accuracy, 93.9\% precision, 96.3\% recall, and a 95.1\% F1-score on the 160-request evaluation set (Section~\ref{sec:eval}). The PE Agent executed enforcement tools on both ALLOW and DENY outcomes. The deterministic rule-based baseline reached 100\% accuracy on the same set, so the benefit of the proposed framework is autonomous decision-making without administrator-authored rules, not higher accuracy.

Each decision carried the agent rationales, the trust score $T$ (when computed), and the identifier of the retrieved policy rule, which makes individual decisions traceable. The core-agent DENY and the fail-safe default ensure that a compromised agent can block access but cannot allow it alone. Reproducibility across repeated runs was not measured and remains a limitation, because LLM inference is non-deterministic.

The main cost is latency. On CPU-only hardware, single-request latency ranged from 45.67\,s to 134.32\,s, compared with 0.7\,ms for the rule-based evaluator (Section~\ref{subsec:latency}). Throughput stayed below 1\,req/min at all tested concurrency levels. The current prototype therefore does not support real-time enforcement.

\subsection{Limitations and Future Research}
The following limitations bound the scope of our findings and point to directions for future work.

\textbf{Single shared model backend.}
All eight agents use one shared local LLM (\texttt{llama3.1} served via Ollama) rather than separate, domain-specialized models. Specialization is achieved only through agent-specific prompts, tool bindings, and isolated ChromaDB collections. Future work should evaluate quantized models per agent under an asynchronous multi-model deployment to determine whether accuracy and latency improve.

\textbf{Adversarial robustness.}
This study does not evaluate direct or indirect prompt injection, jailbreaking, task drift, memory poisoning, or other adversarial-input attacks. Because each agent ingests retrieved policy text and request metadata directly into its prompt, resistance to crafted input and retrieval poisoning remains an open question. A dedicated adversarial stress test is needed to establish robustness empirically.

\textbf{CPU-only inference and latency.}
The testbed runs without GPU acceleration, and the measured latency is likely attributable in large part to this constraint. GPU inference is expected to reduce latency substantially, but this has not been tested. Future work should measure end-to-end latency under GPU inference and with smaller per-agent models.

\textbf{Simulated testbed.}
All results were collected in a simulated five-zone IoBT coalition testbed on a synthetic dataset, not on a live operational network. Behavior under real network conditions, such as packet loss, adversarial jamming, or genuine multi-nation policy conflicts, has not been evaluated.

\section{Conclusion}
\label{sec:conclusion}
This paper proposes Agentic-ZTA, a multi-agent RAG-enabled
framework that operationalizes the NIST SP~800-207 PDP/PEP control
loop through autonomous, domain-specialized LLM agents. Trust
evaluation is decomposed across agents: core agents are evaluated
first, and a DENY from any core agent is enforced directly by the
PEP without waiting for further agents or running trust fusion.
When all core agents return ALLOW, supporting-agent assessments are
combined through confidence-weighted trust fusion, so decisions are
grounded in retrievable policy evidence rather than static rules.

The IoBT testbed demonstrated that the framework can support diverse
access-control outcomes with clear provenance. A compliant
cross-zone write was permitted when all invoked agents converged
above the trust threshold. A compromised-insider read was blocked
directly by a core agent's DENY on cloned-credential evidence,
without invoking trust fusion. An over-privileged cross-zone read
was denied through trust fusion when the combined score fell below
the acceptance threshold. Importantly, when releasability policy
changed in the retrieved knowledge base, the corresponding access
decision changed without retraining or redeployment, showing that
the system can adapt to policy updates at runtime. %These results highlight the value of combining multi-agent orchestration with retrieval-grounded policy reasoning for dynamic Zero Trust enforcement.

%\hl{On the 160-sample access-request evaluation dataset, Agentic-ZTA achieved 95.0\% accuracy, 93.9\% precision, 96.3\% recall, and an F1-score of 95.1\% }\textcolor{red}{, treating DENY as the positive class.}
%\shovan{We define DENY as the positive class because the evaluation goal is to measure how well the system detects and blocks the requests that should be denied, which is the most important outcome in secure access control.}

Overall, the findings suggest that multi-agent LLM orchestration, when
coupled with retrieval-grounded policy knowledge, provides a
practical and transparent mechanism for Zero Trust enforcement in
highly dynamic and adversarial settings.

\section*{Ethics and Privacy Statement}

This work evaluates an access-control decision mechanism on
synthetic and testbed-generated requests, no personal or production organizational data was used. We acknowledge that
any autonomous access-control system carries a risk of erroneous denial affecting legitimate users.

%% JISA / Elsevier typically also expects these as separate
%% declarations near submission (check the Guide for Authors for exact wording and whether they are entered on the portal instead of in the PDF):
 \section*{Declaration of generative AI and AI-assisted technologies in the writing process} 
 During the preparation of this work, the authors have used GPT tools in order to find synonyms for words that are used frequently in the text to avoid repetition and make the manuscript easier to read and more varied in terms of lexicon. After using this tool, the authors reviewed and edited the content as needed and took full responsibility for the content of the publication.
%%   - CRediT author statement
\section*{Declaration of competing interests} 
The authors declare the following financial interests/personal relationships which may be considered as potential competing interests: the authors have patent applications pending, within the same broader research program on ExplainGuard, that are unrelated in scope to the present work. The authors otherwise declare that they have no known competing financial interests or personal relationships that could have appeared to influence the work reported in this paper.

\section*{Data availability statement} 
Access request dataset including access policies is available from the corresponding author upon reasonable request.
\begin{comment}
\section*{Acknowledgements}

This work was supported in part by Tennessee Technological
University.
\end{comment}
%% JISA does not enforce a specific reference style at submission (per the
%% Guide for Authors); elsarticle-num is used to match the original ACM
%% numbered style.
\bibliographystyle{elsarticle-num}
\bibliography{references}

\appendix
\appendix
\section{Use Case 2 Supplementary Data: User, Resource, and Policy Snapshot} This section provides user, resource, access control policies,  access request and normalized access request used for use case2 evaluation. 
\label{sec:appendix-policy-bundle}
\begin{comment}
\subsection{Coalition Zones}
\label{subsec:appendix-zones}

Listing~\ref{lst:appendix-zones} shows the six coalition zones and
their zone-specific enforcement thresholds $\tau_z$, used in the
decision rule of Eq.~\ref{eq:decision}.

\begin{lstlisting}[language=json, caption={Coalition zones and enforcement thresholds.}, label={lst:appendix-zones}]
"zones": [
  {
    "zone_id": "nation_a_tactical",
    "name": "Nation A Tactical",
    "default_threshold": 0.70
  },
  {
    "zone_id": "nation_b_tactical",
    "name": "Nation B Tactical",
    "default_threshold": 0.85
  },
  {
    "zone_id": "mission_hq",
    "name": "Mission HQ",
    "default_threshold": 0.80
  },
  {
    "zone_id": "mission_partner",
    "name": "Mission Partner",
    "default_threshold": 0.85
  },
  {
    "zone_id": "shared_data_fabric",
    "name": "Shared Data Fabric",
    "default_threshold": 0.80
  },
  {
    "zone_id": "classified_network",
    "name": "Classified Network",
    "default_threshold": 0.85
  }
]
\end{lstlisting}
\end{comment}
\subsection{Users}
\label{subsec:appendix-users}

Listing~\ref{lst:appendix-users} shows the subject issuing the
access request evaluated in Section~\ref{sec:uc2}.

\begin{lstlisting}[language=json, caption={User (requesting subject).}, label={lst:appendix-users}]
"users": [
  {
    "user_id": "UAV-045",
    "device_type": "UAV",
    "source_zone": "Nation A Tactical",
    "clearance": "Secret",
    "role": "Tactical UAV",
    "mission_function": "Read C2 server data inside Nation A Tactical",
    "status": "compromised_cloned_credential"
  }
]
\end{lstlisting}

\subsection{Resources}
\label{subsec:appendix-resources}

Listing~\ref{lst:appendix-resources} shows the protected resource
targeted in Section~\ref{sec:uc2}.

\begin{lstlisting}[language=json, caption={resource.}, label={lst:appendix-resources}]
"resources": [
  {
    "resource_id": "RES-C2-SERVER-READ",
    "resource_name": "C2 Server",
    "zone": "Nation A Tactical",
    "classification": "Secret",
    "operations": ["read"]
  }
]
\end{lstlisting}

\subsection{Access Control Policies}
\label{subsec:appendix-policies}

Listing~\ref{lst:appendix-policies} shows the seven policies used for the use case. These policies form RAG knowledge base.  

\begin{lstlisting}[language=json, caption={Access control policy set.}, label={lst:appendix-policies}]
"policy_scope": {
  "source_zone": "Nation A Tactical",
  "destination_zone": "Nation A Tactical",
  "destination_service": "C2 Server",
  "access_mode": "read"
},
"rules": [
  {
    "rule_id": "UC2-RULE-001",
    "effect": "DENY",
    "priority": 100,
    "conditions": {
      "certificate_valid": true,
      "gateway_reported_duplicate_cert": true,
      "concurrent_sessions": {"gte": 2},
      "threat_indicators_any": [
        "DUPLICATE_CERTIFICATE_SESSION",
        "CREDENTIAL_CLONING_SUSPECTED"
      ]
    }
  },
  {
    "rule_id": "UC2-RULE-002",
    "effect": "DENY",
    "priority": 95,
    "conditions": {
      "access_mode": "read",
      "destination_service": "C2 Server",
      "session_metadata": {
        "active_session_ids": {"contains_session_id": true, "min_count": 2}
      },
      "device_credentials": {
        "session_id": {"present_in_active_session_ids": true}
      }
    }
  },
  {
    "rule_id": "UC2-RULE-003",
    "effect": "ALLOW",
    "priority": 50,
    "conditions": {
      "source_zone": "Nation A Tactical",
      "destination_zone": "Nation A Tactical",
      "destination_service": "C2 Server",
      "access_mode": "read",
      "certificate_valid": true,
      "device_credentials": {"clearance": "Secret"},
      "session_metadata": {
        "concurrent_sessions": 1,
        "gateway_reported_duplicate_cert": false
      },
      "threat_indicators_absent": [
        "DUPLICATE_CERTIFICATE_SESSION",
        "CREDENTIAL_CLONING_SUSPECTED"
      ]
    }
  },
  {
    "rule_id": "UC2-RULE-004",
    "effect": "DENY",
    "priority": 90,
    "conditions": {"certificate_valid": false}
  },
  {
    "rule_id": "UC2-RULE-005",
    "effect": "DENY",
    "priority": 85,
    "conditions": {
      "device_credentials": {
        "clearance_any": ["Unclassified", "Confidential"]
      }
    }
  },
  {
    "rule_id": "UC2-RULE-006",
    "effect": "DENY",
    "priority": 80,
    "conditions": {
      "source_device": "172.16.0.45",
      "device_name": "UAV-045",
      "device_identity_mismatch": true
    }
  },
  {
    "rule_id": "UC2-RULE-007",
    "effect": "DENY",
    "priority": 75,
    "conditions": {
      "device_credentials": {"certificate_fingerprint_mismatch": true}
    }
  }
]
\end{lstlisting}

\subsection{Access Request}
\label{subsec:appendix-requests}

\begin{lstlisting}[language=json, caption={Representative access request.}, label={lst:appendix-requests}]
"use_case_access_requests": [
  {
    "use_case": "Use Case 2",
    "request_id": "UC2-UAV-COMPROMISED-READ-0001",
    "access_mode": "read",
    "device_name": "UAV-045",
    "source_zone": "Nation A Tactical",
    "destination_zone": "Nation A Tactical",
    "destination_service": "C2 Server"
  }
]
\end{lstlisting}

\subsection{PEP Normalized Access Request}
\label{subsec:appendix-normalization}

A raw request carries only the minimum fields needed to identify the subject, resource and the requested operation. PEP enriches the request with user/device credentials, session state, threat indicators (if any).

\begin{lstlisting}[language=json, caption={PEP-normalized access request for use case 2.}, label={lst:appendix-normalized-uc2}]
{
  "request_id": "UC2-UAV-COMPROMISED-READ-0001",
  "timestamp": "2026-07-16T10:18:51.752862Z",
  "source_device": "172.16.0.45",
  "device_name": "UAV-045",
  "device_type": "UAV",
  "source_zone": "Nation A Tactical",
  "destination": "172.16.0.20",
  "destination_service": "C2 Server",
  "destination_zone": "Nation A Tactical",
  "access_mode": "read",
  "device_credentials": {
    "certificate": "cert_uav045_valid_signed_unexpired",
    "certificate_valid": true,
    "certificate_fingerprint": "FP-UAV045-ABCD-1234",
    "clearance": "Secret",
    "session_id": "sess_uav045_req_001"
  },
  "session_metadata": {
    "concurrent_sessions": 2,
    "active_session_ids": [
      "sess_uav045_existing_889",
      "sess_uav045_req_001"
    ],
    "gateway_reported_duplicate_cert": true
  },
  "threat_indicators": [
    "DUPLICATE_CERTIFICATE_SESSION",
    "CREDENTIAL_CLONING_SUSPECTED"
  ]
}
\end{lstlisting}
\begin{comment}
\subsection{Selected Agent Response Snapshot}
Listing~\ref{box:uc2_pki} shows the PKI agent's response snapshot.

\begin{tcolorbox}[
  enhanced,sharp corners,breakable,
  colback=respBg,colframe=respSlate,boxrule=0.7pt,
  fonttitle=\bfseries\small,coltitle=white,colbacktitle=respSlate,
  title={Listing~\refstepcounter{promptbox}\thepromptbox\label{box:uc2_pki}: Representative PKI Agent Response Snapshot},
  left=5pt,right=5pt,top=3pt,bottom=3pt,
  fontupper=\footnotesize]
\begin{verbatim}
"PKI": {
  "decision": "DENY",
  "confidence": 0.85,
  "reasoning": [
    {
      "indicator": "certificate_fingerprint_mismatch",
      "description": "Certificate fingerprint mismatch
        detected"
    },
    {
      "indicator": "duplicate_certificate_session",
      "description": "Duplicate certificate session
        detected"
    },
    {
      "indicator": "credential_cloning_suspected",
      "description": "Credential cloning suspected based
        on gateway report"
    }
  ],
  "indicators": [
    "certificate_fingerprint_mismatch",
    "duplicate_certificate_session",
    "credential_cloning_suspected"
  ]
}
\end{verbatim}
\end{tcolorbox}
\end{comment}

\subsection{Agent Decision}
\label{subsec:appendix-uc2-decision-record}

\begin{lstlisting}[language=json, caption={Core-agent decision.}, label={lst:appendix-uc2-decision}]
{
  "request_id": "UC2-UAV-COMPROMISED-READ-0001",
  "core_assessments": [
    {
      "name": "PKI",
      "decision": "ALLOW",
      "confidence": 75,
      "reasoning": [
        "certificate_valid",
        "certificate_fingerprint_match",
        "session_binding_acceptable_one_existing_one_new"
      ]
    },
    {
      "name": "IAM",
      "decision": "DENY",
      "confidence": 80,
      "reasoning": [
        "DUPLICATE_CERTIFICATE_SESSION",
        "CREDENTIAL_CLONING_SUSPECTED"
      ]
    }
  ],
  "matched_rule_id": "UC2-RULE-001",
  "decision": "DENY",
  "denying_agent": "IAM",
  "decision_reason": "Core-agent DENY by IAM."
}
\end{lstlisting}

\end{document}